\documentclass[dvipsnames]{article}
\usepackage[table]{xcolor}
\usepackage{iclr2025_conference,times}

\usepackage{subcaption}
\usepackage{hyperref}
\usepackage{wrapfig}
\usepackage{url}
\usepackage{paralist}
\usepackage{my_commands}
\usepackage{booktabs}
\usepackage{graphicx}

\usepackage[frozencache,cachedir=.]{minted}

\usepackage{tcolorbox}
\usepackage{pifont}
\usepackage{multirow}
\usepackage{listings}

\usepackage{optidef}

\definecolor{darkgreen}{rgb}{0.0, 0.5, 0.0}
\definecolor{citeblue}{rgb}{0.0, 0.3, 0.8}

\hypersetup{
    colorlinks=true,  %
    citecolor=citeblue, 
    linkcolor=citeblue,
    filecolor=citeblue,
    urlcolor=citeblue,
}

\newcommand{\foo}[2]{$\text{#1} \ (+\textcolor{darkgreen}{#2})$}

\title{LiNeS: Post-training Layer Scaling Prevents Forgetting and Enhances Model Merging}

\newcommand{\wiseft}{WiSE-FT\xspace}

\author{Ke Wang\thanks{Equal Contribution} \\
EPFL \\
\texttt{k.wang@epfl.ch}\\
\And
Nikolaos Dimitriadis$\small ^*$\\
EPFL \\
\texttt{nikolaos.dimitriadis@epfl.ch \hspace{9.5pt} }\\
\AND
Alessandro Favero \\
EPFL \\
\texttt{alessandro.favero@epfl.ch}\\
\And
Guillermo Ortiz-Jimenez \\
Google DeepMind \\
\texttt{gortizj@google.com \hspace{58pt} \ } \\
\AND
Fran\c{c}ois Fleuret \\
University of Geneva, Meta FAIR \\
\texttt{francois.fleuret@unige.ch}
\And
Pascal Frossard \\
EPFL \\
\texttt{pascal.frossard@epfl.ch   } \\
}

\iclrfinalcopy %

\usepackage{minitoc}
\begin{document}
\doparttoc %
\faketableofcontents %

\maketitle
\newcommand{\numlayers}{L}
\newcommand{\tvmtlparam}[1]{\ensuremath{\bm{\tau}_{\textrm{MTL}}^{(#1)}}}
\newcommand{\methodname}{\texttt{LiNeS}\xspace}
\newcommand{\algoname}{\texttt{LiNeS}\xspace}
\newcommand{\nummodels}{\ensuremath{N_{\textrm{models}}}}
\newcommand{\ft}[1]{\ensuremath{\bm{\theta}_#1}}
\newcommand{\numparams}{N}
\newcommand{\tvmtl}{\ensuremath{\bm{\tau}_{\textrm{MTL}}}}

\begin{abstract}

    Fine-tuning pre-trained models has become the standard approach to endow them with specialized knowledge, but it poses fundamental challenges. In particular, \textit{(i)} fine-tuning often leads to catastrophic forgetting, where improvements on a target domain degrade generalization on other tasks, and \textit{(ii)} merging fine-tuned checkpoints from disparate tasks can lead to significant performance loss.
    To address these challenges, we introduce LiNeS, Layer-increasing Network Scaling, a post-training editing technique designed to preserve pre-trained generalization while enhancing fine-tuned task performance. LiNeS scales parameter updates linearly based on their layer depth within the network, maintaining shallow layers close to their pre-trained values to preserve general features while allowing deeper layers to retain task-specific representations. In multi-task model merging scenarios, layer-wise scaling of merged parameters reduces negative task interference. LiNeS demonstrates significant improvements in both single-task and multi-task settings across various benchmarks in vision and natural language processing. It mitigates forgetting, enhances out-of-distribution generalization, integrates seamlessly with existing multi-task model merging baselines improving their performance across benchmarks and model sizes, and can boost generalization when merging LLM policies aligned with different rewards via RLHF. Our method is simple to implement, computationally efficient and complementary to many existing techniques. 
    Our source code is available at \href{https://github.com/wang-kee/LiNeS}{github.com/wang-kee/LiNeS}.
\looseness=-1
\end{abstract}

\section{Introduction}
\label{sec:introduction}

Pre-trained models have become the backbone of modern machine learning pipelines \citep{bommasani2021opportunities,Touvron_Martin_Stone_etal_2023}. Their introduction has shifted the paradigm from end-to-end training to fine-tuning \citep{zhuang2020comprehensive}, leading to the proliferation of thousands of fine-tuned checkpoints derived from a few foundation models \citep{rombach2022high,team2023gemini}. To improve downstream performance across multiple tasks or align with multiple preferences \citep{singh2020fusion,matena2022merging,ilharco2023task,yadav2023ties,rame2024rewarded}, model merging techniques combine available checkpoints, avoiding the costly process of joint fine-tuning \citep{ilharco2023task, yadav2023ties}.
However, specializing models introduces trade-offs, such as the forgetting of previously acquired knowledge \citep{aghajanyan2021better} -- a phenomenon known as  \textit{catastrophic forgetting} \citep{mccloskey1989catastrophic}. Furthermore, merging checkpoints fine-tuned on different tasks can lead to significant performance degradation due to task interference \citep{yadav2023ties, wang2024localizing}. \looseness=-1

To mitigate catastrophic forgetting, many works propose regularizing the fine-tuning process \citep{aghajanyan2021better, kumar2022finetuning, gouk2020distance, razdaibiedina2022representation}. Leveraging the insight that shallow layers capture generalizable representations \citep{yosinski2014transferable,neyshabur2020being}, \citet{howard2018universal,dong2022clip} apply lower learning rates to the shallow layers to retain general features.
However, modifying the fine-tuning process can be complex and computationally expensive. This motivates the development of post-training model editing and model merging methods that directly edit the checkpoints in the weight space. For instance, \citet{wortsman2022robust, rame2022diverse} mitigate catastrophic forgetting by interpolating weights between pre-trained and fine-tuned models. In multi-task settings, \citet{yadav2023ties, wang2024localizing} propose methods to reduce interference among tasks when merging multiple checkpoints. Yet, significant performance degradation persists when merging multiple models, leaving this as an open challenge.\looseness=-1

Most model merging methods, however, treat all layers equally, overlooking the earlier insight that shallow layers should remain close to their pre-trained weights to avoid losing the general representations they encode. In this paper, we explore whether this insight can be leveraged post-training. 
We find that reducing the magnitude of shallow-layer updates after fine-tuning can retain single-task performance gains while significantly mitigating forgetting. 

We propose \methodname, \textbf{L}ayer-\textbf{i}ncreasing \textbf{Ne}twork \textbf{S}caling, a post-training, plug-and-play method that directly edits the residual, i.e., the difference between the fine-tuned and pre-trained checkpoint, by applying a scaling coefficient that linearly increases with layer depth.
This scaling effectively preserves the general features captured in the shallow layers of the pre-trained model while retaining task-specific features in the deep layers of the fine-tuned model.
Moreover, we extend \algoname to the multi-task model merging setting, where contributions from one task distort the general features also required by other tasks.
By preserving the general features in the shallow layers, \algoname mitigates task interference and improves multi-task performance.
\looseness=-1

\methodname demonstrates remarkable performance on diverse test scenarios and is orthogonal to existing post-training merging algorithms. It modifies the fine-tuned checkpoint to consistently retrieve nearly full performance on the fine-tuned task while significantly restoring generalization on other tasks. Furthermore, it can be seamlessly integrated with existing weight interpolation methods for improving out-of-distribution generalization \citep{wortsman2022robust}. When merging multiple models, \algoname improves baseline methods for merging checkpoints fine-tuned on multiple tasks in both computer vision and NLP benchmarks \citep{ilharco2023task, yadav2023ties, wang2024localizing} and also enhances performance when merging checkpoints fine-tuned on the same task \citep{wortsman2022model} and merging LLM policies aligned with different rewards \citep{rame2024rewarded} via Reinforcement Learning with Human Feedback (RLHF) \citep{christiano2017deep}.\looseness=-1

Our contributions are as follows:
\vspace{-6pt}
\begin{itemize}
    \item We propose \methodname, a post-training editing technique that preserves the zero-shot generalization of pre-trained models while retaining fine-tuned knowledge by applying layer-wise scaling on parameter updates. For example, in image-classification tasks with CLIP ViT-B/32 checkpoints, 
    \algoname maintains on average 99.8\% of performance on the fine-tuned task while preserving 97.9\% performance of the pre-trained model on other control tasks, effectively mitigating catastrophic forgetting. \looseness=-1
    
    \item We demonstrate that \algoname significantly enhances multi-task model merging baselines, consistently improving performance across benchmarks and architectures in both vision and NLP domains. For instance, we observe a 3.1\% and 4.0\% improvement over Task Arithmetic \citep{ilharco2023task} and Ties-merging \citep{yadav2023ties} respectively, for a 20-task computer vision benchmark with ViT-L/14. \looseness=-1
    
    \item We show that \algoname can be applied to enhance existing weight interpolation methods across various scenarios, improving out-of-distribution generalization, merging multiple checkpoints fine-tuned on the same task with different hyper-parameter configurations, and merging LLM policies aligned with different rewards.\looseness=-1
\end{itemize}

Our proposed method is simple to implement\footnote{PyTorch pseudo-code in \autoref{appendix:pseudocode}. }, orthogonal to many existing approaches, and improves performance in a wide variety of settings.

\section{Related Work}
\label{sec:background}

\paragraph{Representation collapse and regularized fine-tuning} Pre-trained models such as CLIP exhibit strong zero-shot performance across diverse data distributions due to the robust and transferable feature representations learned during pre-training \citep{radford2021learning, jia2021scaling}. However, fine-tuning on specific tasks often harms the zero-shot generalization performance on distributions different from the fine-tuning domain \citep{wortsman2022robust,goyal2023finetune,aghajanyan2021better}. This degradation arises from the distortion of pre-trained features during fine-tuning \citep{kumar2022finetuning}, a phenomenon referred to as \textit{representation collapse} by \citet{aghajanyan2021better}. To mitigate representation collapse, many works have proposed to regularize the fine-tuning process to preserve the general pre-trained features \citep{kumar2022finetuning, goyal2023finetune, gouk2020distance, zhang2022fine, razdaibiedina2022representation, shen2021partial, lee2022surgical}. Some of these approaches take into account that different layers of a model learn distinct features, with the shallower layers capturing more general features and deeper layers specializing in task-specific representations \citep{neyshabur2020being,yosinski2014transferable,adilova2024layerwise}. Specifically, they apply layer-wise learning rate decay, preserving more of the pre-trained features in the shallow layers while allowing deeper layers to specialize for the target domain \citep{clark2020electra, bao2021beit, dong2022clip, howard2018universal, zhang2020revisiting}. However, modifying the fine-tuning process is orders of magnitude more computationally expensive compared to post-training merging methods. \looseness=-1

\paragraph{Weight interpolation and model merging}
\citet{garipov2018loss,draxler2018essentially} showed that two solutions derived from separate training runs can be connected by nonlinear paths of low loss, while \textit{linear mode connectivity} \citep{frankle2020linear} extended the paths to the linear case. 
These insights enabled the transfer of the benefits regarding robustness of (traditional) output ensembles \citep{hansen1990neural,deepensembles} to weight ensembles, reconciling the bias-variance trade-off \citep{belkin2019reconciling} while eliminating the computational cost of multiple inferences \citep{fort2020deep}. These findings can be leveraged to improve performance on single-task \citep{izmailov2018averaging,wortsman2021learning,rame2022diverse,wortsman2022model, jang2024model}, out-of-distribution \citep{wortsman2022robust,rame2023model}, multi-task \citep{ilharco2022patching,dimitriadis2023pareto,dimitriadis2024pareto} and multi-objective alignment  \citep{zhong2024panacea,rame2024warm} settings. Furthermore, model merging can also applied as a scalable approach to unify multiple task-specific models into a single model with multi-task capabilities \citep{ilharco2023task, yadav2023ties}, despite performance loss compared to individual models. 
Several methods have tried to improve multi-task model merging by preserving the important parameters defined via the Fisher Information Matrix \citep{matena2022merging,Tam_Bansal_Raffel_2023}, using heuristics \citep{Davari_Belilovsky_2023,luo2023lcm,jin2023dataless}, 
randomly dropping and rescaling the task vector parameters \citep{yu2024language} or by focusing on resolving weight interference caused by sign disagreements and redundant parameters \citep{yadav2023ties,wang2024localizing}. Recent works use gradient descent to learn the layer-specific merging coefficients per task, e.g., Ada-merging \citep{yang2023adamerging} minimizes entropy in unlabeled test data while aTLAS \citep{zhang2024knowledge} optimizes using cross-entropy loss on validation data. Compared to \methodname, these methods do not incorporate any prior knowledge on early vs. deep layers and require training, resulting in significant computational overheads. 
\looseness=-1

\section{Post-training Layer-wise Scaling Mitigates Forgetting}
\label{sec:motivation}

In this section, we present the key insight of our work: Scaling down the updates of shallow layers after fine-tuning can mitigate catastrophic forgetting and restore zero-shot generalization while preserving performance on the target task.

\paragraph{Notation }  We consider a pre-trained model $\ptm\in\r^\numparams$ with $\numparams$ parameters. Fine-tuning on a specific task $t$ results in the fine-tuned weights $\ft{t}$. The difference between these two sets of weights, $\tv{t}=\ft{t}-\ptm$, is referred to as the \textit{task vector} or \textit{residual} for task $t$ \citep{ilharco2023task} and represents the updates made during fine-tuning.

\paragraph{Fine-tuning leads to catastrophic forgetting} 
We quantitatively demonstrate the phenomenon of catastrophic forgetting with the following experiments. Consider the 8-task image classification benchmark studied in \cite{ilharco2023task}. We fine-tune a CLIP ViT-B/32 model on each task, measuring performance on the fine-tuned task -- referred to as the \textit{target task} -- and the remaining 7 tasks -- the \textit{control tasks}. %
The averaged results over all target and control task combinations, shown in \autoref{tab:fine-tuning}, demonstrate that while fine-tuning significantly improves accuracy on the target task, it drastically reduces accuracy on the control tasks, underscoring the loss of the model's zero-shot generalization abilities. 
\looseness=-1

\begin{wraptable}[11]{r}{0.49\textwidth} 
\vspace{-12pt}
\centering
\caption{Fine-tuning harms generalization on control tasks. Our proposed post-training edition leads to a superior trade-off between performance on target and control tasks.\looseness=-1}
\label{tab:fine-tuning}
\vspace{-4pt}  %
\resizebox{\linewidth}{!}{%
\begin{tabular}{@{}ccc@{}}
\toprule
Model / Accuracy & Target & Control  \\ \midrule
Pre-trained  & 48.3 & 48.3 \\
Fine-tuned  & 90.5 & 38.0 \\ 
Fine-tuned+\algoname (ours)  & 90.3 & 48.0 \\ 
\bottomrule
\end{tabular}
}
\end{wraptable}

\begin{figure}[t]
    \centering
    \includegraphics[width=\linewidth]{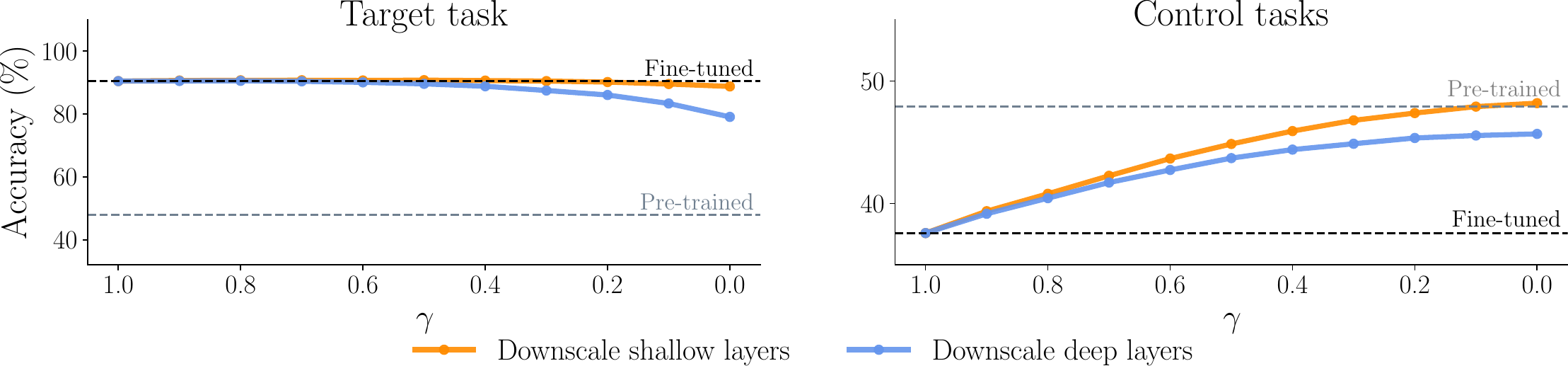}
    \caption{Downscaling the shallow layers maintains the fine-tuned performance on target tasks (orange line, left), while restoring zero-shot performance from pre-trained model on control tasks (orange line, right). The performance for downscaling deep layers instead is presented in blue lines, which underperforms downscaling shallow layers in both cases. $\gamma$ represents the minimum scaling factor applied to the layers, where a smaller $\gamma$ leads to stronger downscaling strength, with $\gamma = 1$ restoring the original fine-tuned model.
    \looseness=-1}
    \label{fig:downscale_id_ood}
\end{figure}

\paragraph{Shallow-layer updates impact minimally on target task accuracy} Most parameter updates during the fine-tuning process are redundant, as similar performance is achievable without updating most pre-trained weights \citep{yadav2023ties, wang2024localizing, he2024localize}. 
Moreover, prior work shows that task-specific features are often concentrated in deeper layers of the network \citep{neyshabur2020being,yosinski2014transferable, raghu2019transfusion}. Based on these observations, we hypothesize that updates to the shallow layers contribute minimally to target tasks. \looseness=-1
To test this, we progressively downscale the updates to shallow layers after fine-tuning. 
Specifically, we apply a scaling factor to the updates to the $\ell$-th layer $\bm{\tau}^{(\ell)}$, defined as: 
$
    \lambda^{(\ell)}= \gamma + (1-\gamma) \frac{\ell-1}{\numlayers-1}, \ \forall \ell \in[\numlayers], \gamma\in[0,1],
$
This linearly scales the updates from a factor of $\gamma$ for the first layer to $1$ for the last one. 
As a result, fine-tuning updates to the shallow layers are scaled down more aggressively, with later layers experiencing progressively smaller reductions.  We then reintroduce the scaled task vector into the pre-trained model and measure its performance on the fine-tuned task. 
\autoref{fig:downscale_id_ood} (left) shows the results of this experiment for the CLIP ViT-B/32 checkpoint fine-tuned across the 8 tasks, where $\gamma$ is progressively decreased to strengthen the downscaling effect. We observe that, even with strong downscaling of shallow layers, the target task accuracy remains nearly unaffected. In contrast, when we downscale the deeper layers, target task accuracy drops significantly.
These results support our hypothesis that shallow-layer updates are largely unnecessary for maintaining accuracy on the target task.\looseness=-1
\looseness=-1

\paragraph{Shallow-layer updates undermine zero-shot generalization}

While shallow-layer updates have minimal impact on target-task accuracy, they distort the general features learned during pre-training, which reside primarily in the shallow layers \citep{neyshabur2020being, yosinski2014transferable, raghu2019transfusion}. We hypothesize that the degradation of performance on control tasks is largely due to these distortions in the shallow layers. 
Using the same experimental setup, we now evaluate the zero-shot performance on the control tasks, i.e., the other 7 unseen tasks. As shown in \autoref{fig:downscale_id_ood} (right), as the strength of the shallow-layer downscaling increases, the accuracy on control tasks approaches the original pre-trained model’s performance. This shows that by reducing the shallow-layer updates, we can restore most of the zero-shot performance that is lost during fine-tuning. \looseness=-1

\begin{wrapfigure}[20]{r}{0.5 \textwidth} %
    \centering
    \includegraphics[width=1.0\linewidth]{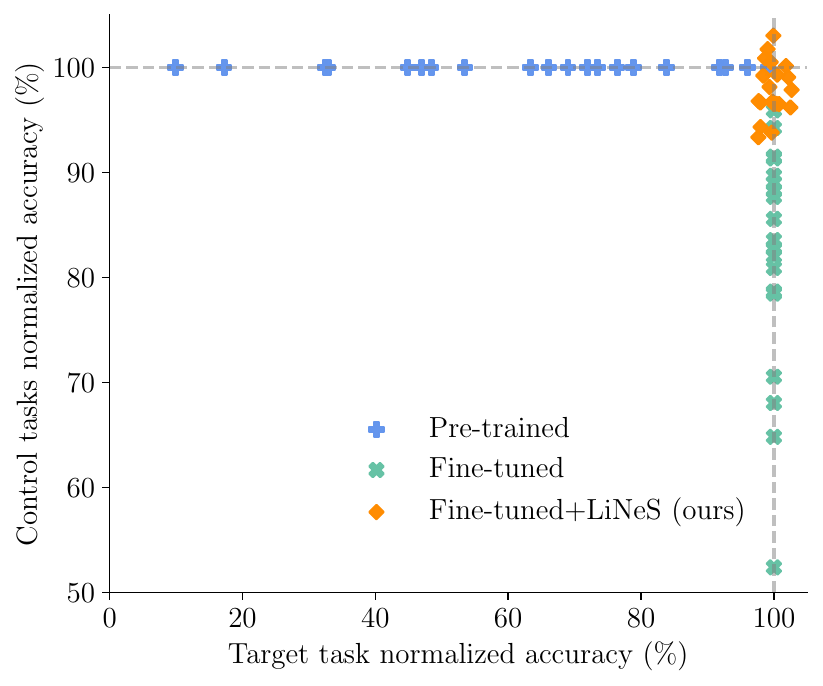}
        \caption{Our linear scaling (\algoname) retains performance on both control and fine-tuned target tasks. Each dot represents a different model.\looseness=-1}
    \label{fig:id_ood_norm_acc}
\end{wrapfigure}

\paragraph{Improved trade-off between target and control performance} 
To optimize the trade-off between target and control task performance, we select a scaling coefficient $\gamma$ for each model
that maximizes a weighted balance between these two objectives, as detailed in Appendix~\ref{app:trade_off_selection}.
After selecting the optimal scaling coefficient, the test results are shown in the final row of \autoref{tab:fine-tuning}. 
Our post-training method preserves target task accuracy with a minimal 0.2\% difference while improving control task performance by 10\%, compared to the fine-tuned model.

We further apply the same method to a 20-task computer vision benchmark \citep{wang2024localizing}. \looseness=-1
For evaluation, we report both the \textit{target task normalized accuracy} and the \textit{control task normalized accuracy} on the 19 tasks, where accuracy is normalized by the performance of the fine-tuned model for the target task and the zero-shot accuracy of the pre-trained model for the control tasks. 
We compare to fine-tuned models on each task and the pre-trained model as baselines. \autoref{fig:id_ood_norm_acc} shows that fine-tuning degrades zero-shot generalization, as indicated by the performance drop on control tasks. In contrast, our post-training scaling method significantly improves generalization while maintaining near-full target task accuracy. On average, our method achieves a target task normalized accuracy of 99.8\% and a control task normalized accuracy 97.9\%. This demonstrates its effectiveness in preserving both task-specific knowledge from fine-tuned checkpoints and the generalization capabilities of the pre-trained model. The full breakdown of results by task is available in \autoref{fig:id_ood_norm_acc_app} in Appendix.\looseness=-1

In Appendix, we show that catastrophic forgetting happens with models fine-tuned with LoRA \citep{hu2022lora} as well. As shown in \autoref{tab:lora ft}, higher expressivity in the form of higher ranks increases target accuracy for LoRA but at the cost of lower performance on control tasks. Still, \algoname significantly improves control performance while minimally affecting target accuracy. Furthermore, in \autoref{fig:convnext} of the appendix, we show that similar benefits can be observed for convolutional architectures such as ConvNeXt \citep{liu2022convnet}. 
Finally, we provide a performance comparison between editing models with \algoname and regularized-fine-tuning-based methods in Appendix~\ref{app:regularized_finetuning}, including applying different learning rates per layer. Also in these cases, \algoname demonstrates superior performance on control tasks, while being much more computationally efficient. \looseness=-1

\section{Method}
\label{sec:method}

Motivated by the results of the previous section for mitigating forgetting, we propose \algoname for \textbf{L}ayer-\textbf{i}ncreasing \textbf{Ne}twork \textbf{S}caling, a simple post-training technique that linearly rescales the updates of different layers in the task vector based on their depth in the network. \algoname is designed to retain general features in the shallow layers while preserving the task-specific adaptations in the deeper layers. \looseness=-1

Given a task vector $\bm{\tau}$ with $L$ layer blocks
we apply the layer-wise linear scaling to adjust the contributions of shallow and deep layers using the following formulation:
\begin{equation}
\label{eq:linear_scaling_a_b}
\bm{\tau}_{\textrm{LiNeS}} = \texttt{concat}\left( \lambda^{(1)} \bm{\tau}_{\textrm{}}^{(1)}, \dots, \lambda^{(\numlayers)} \bm{\tau}_{\textrm{}}^{(\numlayers)} \right), \quad \text{where} \ \ \lambda^{(\ell)} = \alpha + \beta \frac{\ell-1}{\numlayers-1} , \quad \forall \ell \in [\numlayers].
\end{equation}
As a result, the layers in $\bm{\tau}$ are progressively scaled with a factor between $\alpha$ for the first layer and $\alpha+\beta$ for the last layer, with intermediate layers scaled with a linearly increasing schedule depending on their depth. The final model $\bm\theta$ is then obtained by summing the pre-trained model weights and the edited task vector, i.e., $\bm\theta = \ptm + \bm{\tau}_{\text{LiNeS}}$. Notice that, in \autoref{eq:linear_scaling_a_b}, $\bm{\tau}$ can correspond to either a single-task residual or, in the context of model merging, a multi-task vector obtained by merging the residuals of multiple checkpoints fine-tuned starting from a common initialization. Additional details on this process are provided in the next section.
Setting $\alpha=\beta=0$ corresponds to the pre-trained model, while $\alpha=1, \beta=0$ is the fine-tuned model in the case that $\bm{\tau}$ is a single-task vector. \looseness=-1

In practice, we find that tuning just one hyper-parameter (either $\alpha$ or $\beta$) is often sufficient to achieve a good balance between target task performance and generalization. Specific details on hyper-parameter tuning for different applications are provided in the experimental sections.
The linear scaling method introduced in Section~\ref{sec:motivation} corresponds to \algoname by setting $\alpha=\gamma$ and $\beta=1-\gamma$.  This formulation generalizes our previous approach, offering a flexible way to adjust the contributions of different layers based on the task requirements.
\looseness=-1

\section{Model Merging Experiments}
\label{sec:experiments}

We empirically verify the effectiveness of applying \algoname across diverse application domains. 
\autoref{sec:robust_ft} presents results for improving robust fine-tuning \citep{wortsman2022robust} for OOD generalization; 
\autoref{sec:multi-task model merging} focuses on  improving existing multi-task merging methods \citep{ilharco2023task,yadav2023ties,wang2024localizing} in both vision and NLP benchmarks. In \autoref{sec:single-task model merging}, we apply \algoname and improve the merging of single-task fine-tuned models within the setting of Model Soups \citep{wortsman2022model}, and finally, we enhance merging foundation models fine-tuned on different rewards \citep{rame2024rewarded} in \autoref{sec:rewarded_soup}. \looseness=-1

\subsection{Improving Robust Fine-tuning for OOD Generalization}
\label{sec:robust_ft}

\begin{figure}[t]
    \centering
    \includegraphics[width=1\linewidth]{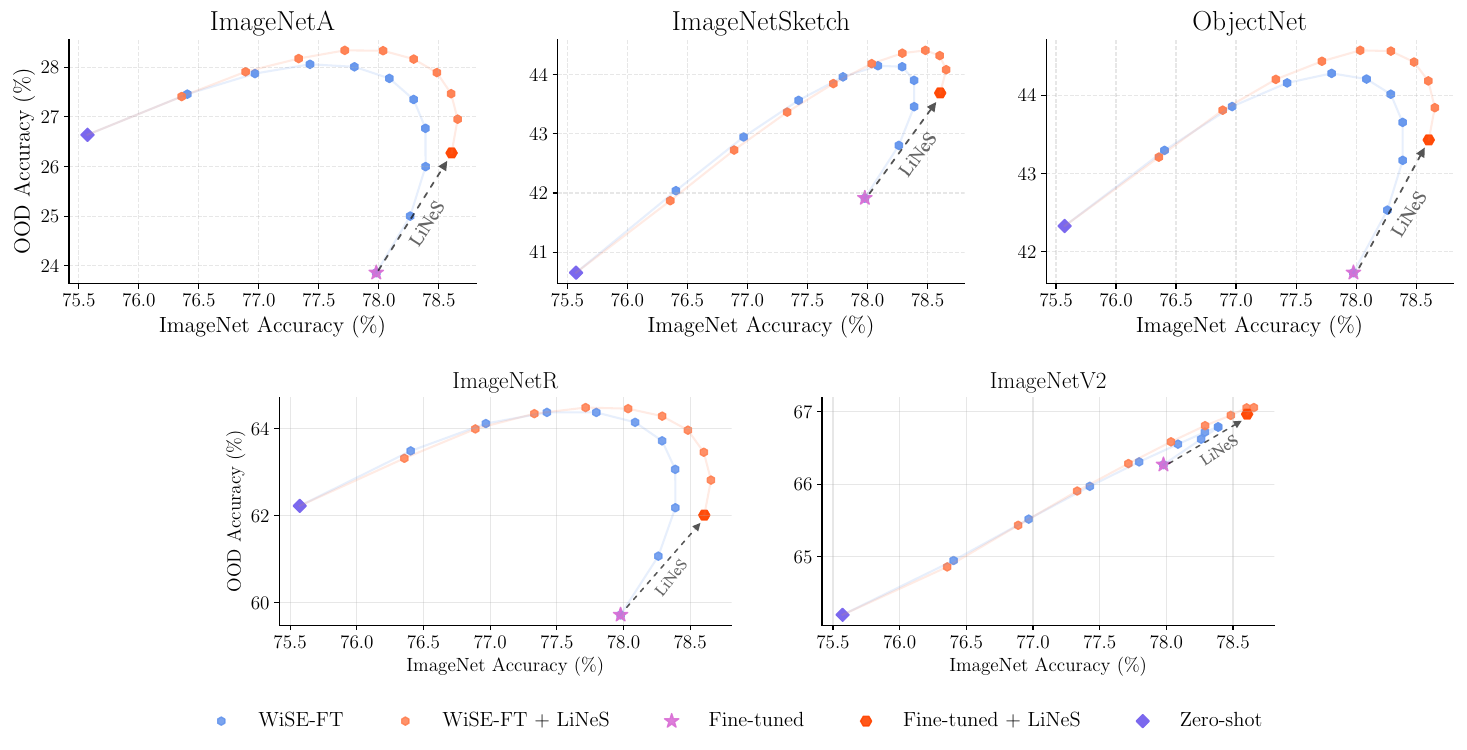}
    \caption{Application of \algoname to \wiseft \citep{wortsman2022robust} improves performance on ImageNet and five different distribution shifts, resulting in a dominating Pareto Front over \wiseft. \looseness=-1}
    \label{fig:robust_fine-tuning}
\end{figure}

We first consider the setting of robust fine-tuning or \wiseft \citep{wortsman2022robust}, where linearly interpolating between the pre-trained and the fine-tuned weights improves model performance on OOD datasets. The interpolation is equivalent to scaling the residual $\bm\tau$: $(1-\gamma)\bm\theta_0 + \gamma\bm\theta = \bm\theta_0 + \gamma \bm\tau$, for  $\gamma\in[0, 1]$. %
We apply \algoname to the residual $\bm\tau$.
Following \cite{wortsman2022robust}, we evaluate CLIP models fine-tuned on ImageNet \citep{deng2009imagenet}, considering 5 OOD datasets, namely ImageNetSketch \citep{wang2019learning},
ImageNet-A \citep{hendrycks2021natural},
ImageNet-R \citep{hendrycks2020many}, ObjectNet \citep{objectnet}, ImageNet-V2 \citep{recht2019imagenet}. 
\looseness=-1

We apply this \methodname to each of the 70 fine-tuned checkpoints\footnote{The checkpoints are CLIP ViT-B/32 models fine-tuned on ImageNet with different hyper-parameters.} provided by \cite{wortsman2022model} setting $\alpha=\beta=0.5$. We present the average results in \autoref{fig:robust_fine-tuning}, comparing the performance of \wiseft with and without applying \algoname on the 5 OOD datasets. 
Without applying \wiseft, \algoname already enhances both the ID and OOD performance of the fine-tuned models by a notable margin. Starting from this edited model and applying the \wiseft interpolation with the pre-trained weights leads to a Pareto Front \citep{Caruana_1997} that consistently dominates the one by \wiseft across all distribution shifts, illustrating the applicability of the proposed method across various distribution shifts. A granular result for applying \algoname to each of the 70 checkpoints is provided in Appendix~\ref{app:wiseft_individual}, further highlighting its universal effectiveness across models. 
We also report similar findings in \autoref{fig:wiseft_vit_b_16} in Appendix for a CLIP ViT-B/16 checkpoint fine-tuned on ImageNet, using the same hyper-parameters as \cite{wortsman2022robust}.
\looseness=-1
    
\subsection{Improving Multi-task Model Merging}
\label{sec:multi-task model merging}

\begin{table}[t]
\caption{Results for multi-task model merging in vision classification benchmarks of 8 tasks \citep{ilharco2023task}, 14 tasks, and 20 tasks \citep{wang2024localizing} for different vision transformer architectures. Applying \algoname improves baseline performance for all benchmark/architecture combinations.\looseness=-1}
\vspace{-5pt} 
\label{tab:vision}
\resizebox{\textwidth}{!}{%

\begin{tabular}{@{}lccclllllll@{}}
\toprule
\multirow{2}{*}{Method} & \multirow{2}{*}{\begin{tabular}[c]{@{}c@{}}with \\ \algoname\end{tabular}} &  &  & \multicolumn{3}{c}{ViT-B/32} & \multicolumn{1}{c}{} & \multicolumn{3}{c}{ViT-L/14} \\ \cmidrule(lr){5-7} \cmidrule(l){9-11} 
 &  &  &  & 8 tasks & 14 tasks & 20 tasks &  & 8 tasks & 14 tasks & 20 tasks \\ \midrule
\rowcolor[HTML]{EFEFEF} Zero-shot &  &  &  & 48.3 & 57.3 & 56.1 &  & 64.8 & 68.3 & 65.3 \\
\rowcolor[HTML]{EFEFEF} Fine-tuned &  &  &  & 90.5 & 89.5 & 90.4 &  & 94.0 & 93.3 & 94.0 \\ \midrule
\multirow{2}{*}{Task Arithmetic} & \ding{55} &  &  & 69.7 & 65.0 & 60.3 &  & 84.0 & 79.2 & 74.0 \\
 & \ding{51} &  &  & \textbf{\foo{74.2}{4.5}} & \textbf{\foo{69.1}{4.1}} & \textbf{\foo{63.4}{3.1}} &  & \textbf{\foo{86.5}{2.5}} & \textbf{\foo{82.2}{3.0}} & \textbf{\foo{77.1}{3.1}} \\ \midrule
\multirow{2}{*}{Ties-Merging} & \ding{55} &  &  & 73.6 & 67.6 & 63.1 &  & 85.6 & 79.3 & 75.6 \\
 & \ding{51} &  &  & \textbf{\foo{77.2}{3.6}} & \textbf{\foo{72.1}{4.5}} & \textbf{\foo{67.2}{4.1}} &  & \textbf{\foo{88.0}{2.4}} & \textbf{\foo{82.5}{3.2}} & \textbf{\foo{79.6}{4.0}} \\ \midrule
\multirow{2}{*}{Consensus Merging} & \ding{55} &  &  & 74.5 & 70.1 & 65.3 &  & 85.2 & 81.9 & 78.7 \\
 & \ding{51} &  &  & \textbf{\foo{77.6}{3.1}} & \textbf{\foo{73.6}{3.5}} & \textbf{\foo{68.6}{3.3}} &  & \textbf{\foo{87.3}{2.1}} & \textbf{\foo{84.0}{2.1}} & \textbf{\foo{81.0}{2.3}} \\ \bottomrule
\end{tabular}
}
\end{table}

In this section, we extend \algoname to improve multi-task merging algorithms, aiming to combine multiple models fine-tuned independently on different tasks into a single model \citep{matena2022merging,ilharco2023task, ortiz2023task, yadav2023ties,hazimeh2024task}. Task arithmetic \citep{ilharco2023task} proposed to decouple the contributions of the pre-trained model and individual task vectors, first generating a multi-task vector $\tvmtl=g(\tv{1}, \dots, \tv{\numtasks})$ with a merging function $g: \r^N\times\cdots\times\r^N \mapsto \r^N$, and then adding back to the pre-trained checkpoint with a scaling factor to construct a multi-task model $\bm{\theta}=\ptm+\lambda \cdot \tvmtl$. The scalar coefficient $\lambda$ is tuned using a held-out validation set. Recent works \citep{yadav2023ties, wang2024localizing} follow the same protocol while improving the merging function $g$ for retaining more task information. We refer to Appendix~\ref{app:baselines} for a more detailed explanation of these methods.
\looseness=-1

However, significant performance loss occurs between the merged multi-task model and the original fine-tuned checkpoints. This performance decrease partially stems from interference \citep{yadav2023ties,wang2024localizing} among task vectors, where the contribution of one task negatively impacts performance on others, leading to overall degradation. 
Task interference is linked to catastrophic forgetting, as the individual task vectors lose a significant amount of generalization ability to other tasks after fine-tuning and merging them leads to interference among each other. Therefore, we can edit each task vector with \algoname before merging to restore the generalization to other tasks, or for simplicity, edit directly the merged multi-task vector to preserve the shallow and general features that are beneficial across tasks. 

We enhance the merging methods by applying \algoname on the merged multi-task vector $\tvmtl$.
For the linear scaling schedule, we tune only $\beta$ and set $\alpha$ using a heuristic that adjusts based on both the number of merged models and the merging method. Specifically, for task arithmetic which aggregates the individual task vectors through a simple summation operation: $\bm\tau_\text{sum} = \sum_{i=1}^{\nummodels} \bm{\tau}_i$, we set $\alpha = 1/\nummodels$. For other merging strategies which result in $\tvmtl$ with different magnitudes of norms, e.g., aggregation with summation leads to a norm $\times\nummodels$ larger compared to averaging, we further multiply by a scaling term $\left\| \bm{\tau}_{\text{sum}} \right\| / \|\tvmtl\|$ to normalize their norm to simple summation. Overall, we set the intercept $\alpha$ to: \looseness=-1
\begin{equation}
    \label{eq:heuristic}
    \alpha =   \frac{1}{\nummodels}  \frac{\left\| \bm{\tau}_{\text{sum}} \right\|}{\|\tvmtl\|}, \textrm{ where } \bm\tau_{\text{sum}} = \sum_{i=1}^{\nummodels} {\bm{\tau}_i}
\end{equation}
Therefore, we only tune $\beta$ and search over the same range as the constant scaling $\lambda$ used by the aforementioned merging techniques. As a result, \algoname shares the same computational requirements as the baseline merging methods; we provide more details for the hyper-parameters in Appendix~\ref{app:hyper_multi_task}, as well as a sensitivity analysis to the hyper-parameters in Appendix~\ref{app:sensitivity_analysis}.
Specifically, we consider various multi-task model merging baselines, namely Task Arithmetic \citep{ilharco2023task}, Ties-merging \citep{yadav2023ties}, Consensus Merging \citep{wang2024localizing}, enhancing them with \algoname and evaluate on both computer vision and NLP benchmarks.
\looseness=-1

\subsubsection{Computer vision}

We experiment with the 8-task image classification benchmark proposed by \cite{ilharco2023task}, as well as the more challenging 14-task and 20-task benchmarks from \cite{wang2024localizing}. Detailed descriptions of task composition appear in Appendix~\ref{app:benchmark contents}. We also examine the efficacy of \algoname across the model scale axis, studying three vision transformer  \citep{dosovitskiy2021an}, namely ViT-B/32, ViT-B/16 and ViT-L/14, as CLIP visual encoders \citep{radford2021learning} . 

\autoref{tab:vision} presents the results for ViT-B/32 and ViT-L/14, while Appendix~\ref{app:vit_b_16} contains the ViT-B/16 experiments. We observe that \algoname provides a significant improvement to \textit{all} baseline merging methods across \textit{all} tested scenarios, regardless of model sizes and total number of tasks. For example, for the 8-task benchmark with ViT-B/32, \algoname improves task arithmetic by 4.5\%, Ties-merging by 3.6\% and consensus merging by 3.1\%. For the challenging 20-task benchmark with ViT-L/14, \algoname leads to consistent and significant improvements, improving task arithmetic by 3.1\%, Ties-merging by 4.0\% and consensus merging by 2.3\%. The detailed performance on individual tasks for each tested scenario is presented in Appendix~\ref{app:individual_performance}.

\subsubsection{Natural Language Processing}
\label{sec:nlp with t5}

We also evaluate the effectiveness of \algoname in NLP domain, including a 7-task NLP benchmark \citep{yadav2023ties}, an 8-task Question-Answering benchmark  \citep{zhou2022not}, and their combined 11-task benchmark \citep{wang2024localizing}.  Appendix~\ref{appendix:experimental details} details the experimental settings. Following \cite{Tam_Bansal_Raffel_2023}, we adopt a variant of T5-large model \citep{colin2020exploring}, namely T5-large-LM-Adapt \citep{lester-etal-2021-power}, and use their provided checkpoints. While T5-large contains both encoder and decoder networks, we apply \algoname only to the decoder, as our findings in Appendix~\ref{app:shallow_t5} indicate that applying the edition to the decoder leads to similar observations to vision. \looseness=-1

The performance of applying \algoname to baseline methods with T5-large across various NLP tasks is summarized in \autoref{tab:nlp with t5}. \algoname consistently improves multi-task performance across baseline merging methods and benchmarks with a notable margin. For example, on the 7 NLP tasks benchmark, \algoname improves task arithmetic by 4.5 points, and consensus merging by 1.9 points. Meanwhile, \algoname outperforms Ties-merging by 3.0\% and 2.4\% for the 8-QA benchmark and 11-NLP benchmark, respectively. \looseness=-1

\begin{table}[t!]
\caption{Results for multi-task model merging methods in three NLP benchmarks with T5-large model. \algoname improves baseline performance across merging methods and benchmarks.\looseness=-1}
\vspace{-5pt} 

\label{tab:nlp with t5}
\centering
\resizebox{\textwidth}{!}{%
\begin{tabular}{@{}lclll@{}}
\toprule
\multirow{2}{*}{Method} & \multirow{2}{*}{\begin{tabular}[c]{@{}c@{}}with \\ \algoname\end{tabular}} & \multicolumn{3}{c}{T5-large \citep{lester-etal-2021-power}} \\ \cmidrule(l){3-5} 
 &  & \multicolumn{1}{c}{7 NLP tasks \citep{yadav2023ties}} & \multicolumn{1}{c}{8 QA tasks \citep{zhou2022not}} & \multicolumn{1}{c}{11 NLP tasks \citep{wang2024localizing}} \\ \midrule
\rowcolor[HTML]{EFEFEF} Zero-shot & \multicolumn{1}{l}{} & 44.9 & 33.1 & 36.9 \\
\rowcolor[HTML]{EFEFEF} Fine-tuned &  & 85.9 & 80.7 & 78.7 \\ \midrule
\multirow{2}{*}{Task Arithmetic} & \ding{55} & 71.9 & 63.8 & 63.6 \\
 & \ding{51} & \textbf{\foo{76.4}{4.5}} & \textbf{\foo{67.6}{3.8}} & \textbf{\foo{66.2}{2.6}} \\ \midrule
\multirow{2}{*}{Ties-Merging} & \ding{55} & 71.6 & 63.0 & 64.0 \\
 & \ding{51} & \textbf{\foo{72.0}{0.4}} & \textbf{\foo{66.0}{3.0}} & \textbf{\foo{66.4}{2.4}} \\ \midrule
\multirow{2}{*}{Consensus Merging} & \ding{55} & 73.5 & 68.6 & \textbf{67.5} \\
 & \ding{51} & \textbf{\foo{75.4}{1.9}} & \textbf{\foo{69.3}{0.7}} & \textbf{\foo{67.5}{0.0}} \\ \bottomrule
\end{tabular}

}
\end{table}

\subsection{Improving Model Soups for Merging single-task models}
\label{sec:single-task model merging}

Averaging in weight space multiple models fine-tuned on the same task derived from the same pre-trained model has been shown to increase target performance \citep{wortsman2022model,rame2022diverse}. In this section,  we investigate whether \algoname can enhance the test performance when merging single-task models.

We follow the setting in Model Soups \citep{wortsman2022model} and merge 70 CLIP ViT-B/32 checkpoints fine-tuned on ImageNet \citep{deng2009imagenet} using different hyper-parameters, plus the pre-trained checkpoint. We consider both variants introduced in \citet{wortsman2022model}, namely uniform and greedy soup. We refer to Appendix~\ref{app:experimental_details_single_task} for details regarding these methods and experimental settings.
For both cases, the weight-averaging process can be decomposed as follows:
\begin{equation}
\vspace{-2pt}
\bm{\theta}_{\text{soup}} = 
\bm{\theta}_0 + \bm{\tau}_{\text{soup}}, \ \text{where} \ \bm{\tau}_{\text{soup}} = \frac{1}{\nummodels} \sum_{i=1}^{\nummodels} {(\bm\theta_i - \bm\theta_0)}
\label{eq:uniform_soup}
\end{equation}

We apply \algoname to $\bm{\tau}_{\text{soup}}$  fixing $\alpha=1$ and searching over $\beta$. As a baseline, we also consider task arithmetic, where we search for a constant scaling factor on $\bm{\tau}_{\text{soup}}$. Note that both settings introduce one hyper-parameter to vanilla model soups, and refer to Appendix~\ref{app:experimental_details_single_task} for a detailed description of the modifications. \autoref{tab:model_soups} summarizes the results and shows that \algoname improves over vanilla soups and task arithmetic for both uniform and greedy soup by $0.48\%$ and $0.15\%$ on ImageNet, respectively. 
We report the best-performing model and the average performance as baselines.
Finally, our proposed method compounds the gains from the greedy soup and leads to the best-performing model. 
\looseness=-1

\subsection{Improving Rewarded Soups}
\label{sec:rewarded_soup}

In this section, we explore the effectiveness of \methodname for merging foundation models fine-tuned on different rewards. We consider the Rewarded Soups setting \citep{rame2023model}, which interpolates the weights $\bm{\theta}_1$ and $\bm{\theta}_2$ of two LLM policies, each optimized for a distinct reward $R_1$ and $R_2$, respectively. \looseness=-1

\looseness=-1 Starting with an LLM parameterized by weights $\bm{\theta}_0$, we first fine-tune it using supervised fine-tuning (SFT) on labeled demonstrations. 
From the resulting weights $\bm{\theta}_{\rm SFT}$, we then apply Reinforcement Learning from Human Feedback (RLHF) \citep{christiano2017deep,ouyang2022training}, training two independent policies via Proximal Policy Optimization (PPO) \citep{schulman2017proximal} to maximize the rewards $R_1$ and $R_2$ respectively. To merge these policies, we linearly interpolate the residuals $\bm{\tau}_1=\bm{\theta}_1-\bm{\theta}_{\rm SFT}$ and $\bm{\tau}_2=\bm{\theta}_2-\bm{\theta}_{\rm SFT}$, defining a continuous set of rewarded policies: \looseness=-1
\begin{equation}
    \bm{\theta}_{RS} = \bm{\theta}_{\rm SFT} + \lambda \bm{\tau}_1 + (1-\lambda) \bm{\tau}_2, \quad \lambda \in [0,1],
\end{equation}
where the coefficient $\lambda$ models the user's preferences. 
We apply \algoname to the weighted-sum residual: $\lambda \bm{\tau}_1 + (1-\lambda) \bm{\tau}_2$, fixing $\alpha=\beta=1$ for computational reasons.

\begin{table}[t]
    \centering
    \begin{minipage}{0.51\textwidth} %
        \centering
        \caption{\algoname improves performance over Model Soups \citep{wortsman2022model}, for both uniform and greedy soup in merging multiple checkpoints fine-tuned on ImageNet with different hyper-parameter configurations. \looseness=-1}
        \vspace{6pt}
        \resizebox{\linewidth}{!}{%
            \begin{tabular}{@{}lll@{}}
                \toprule
                Method & Enhancements & ImageNet Acc. \\ \midrule
                \rowcolor[HTML]{EFEFEF} Averaged accuracy & / & 77.98 \\ 
                \rowcolor[HTML]{EFEFEF} Best individual model & / & 80.36 \\ \midrule
                \multirow{3}{*}{Uniform soup} & / & 79.99 \\
                & Task Arithmetic & 80.17 \\
                & \algoname & \textbf{\foo{80.47}{0.48}} \\ \midrule
                \multirow{3}{*}{Greedy soup} & / & 81.01 \\
                & Task Arithmetic & 81.01 \\
                & \algoname & \textbf{\foo{81.16}{0.15}} \\ \bottomrule
            \end{tabular}
        }
        \label{tab:model_soups}
    \end{minipage}
    \hfill
    \begin{minipage}{0.45\textwidth} %
        \centering
        \includegraphics[width=1.0\linewidth]{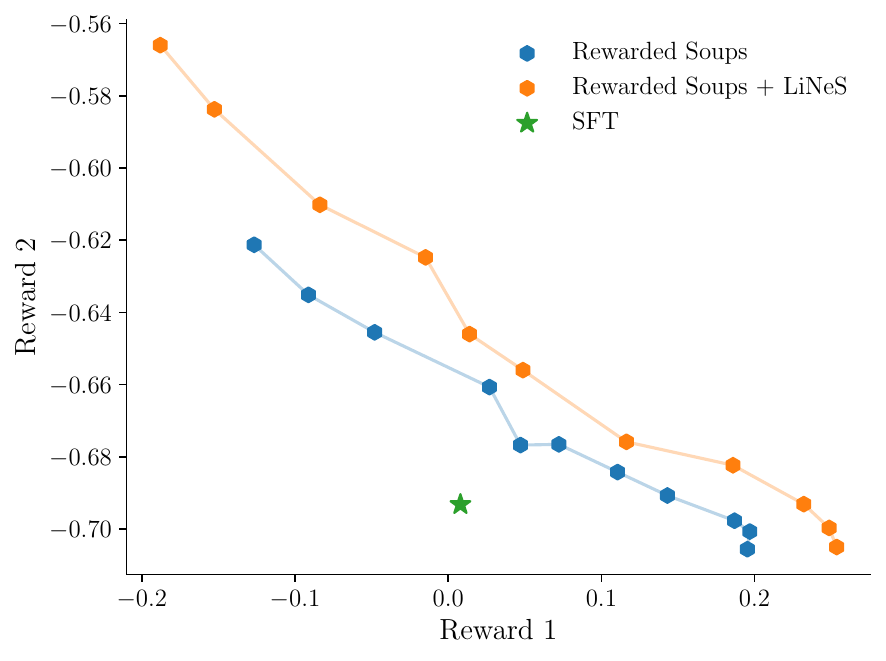}
        \vspace{-18pt}
        \captionof{figure}{Applying \methodname to Rewarded Soups \citep{rame2023model} improves merging of LLM policies RL fine-tuned on different rewards with a dominating Pareto Front. \looseness=-1}
        \label{fig:rewarded}
    \end{minipage}
\end{table}

In our experiment, we use LLaMA-2 7B \citep{Touvron_Martin_Stone_etal_2023} and the Reddit Summary task \citep{stiennon2020learning}, which consists of 14.9k post-summary pairs. We fine-tune the model using LoRA \citep{hu2022lora} with $r_{\rm \scriptscriptstyle LoRA}=64$, $\alpha_{\rm \scriptscriptstyle LoRA}=128$, and 0.05 dropout. We employ two reward models: GPT2-reward-summarization -- which scores summaries based on human preferences -- and BART-faithful-summary-detector \citep{CZSR21} -- which evaluates the faithfulness of the generated summary to the source post.  To evaluate the models, we use a subset of 1k samples from the test set, generate the responses, and compute the average score for each reward dimension.

In \autoref{fig:rewarded}, we present the empirical Pareto Fronts for both Rewarded Soups and Rewarded Soups+\methodname. \methodname consistently outperforms the vanilla Rewarded Soups across the full preference space, Pareto dominating the baseline. This result highlights the generality of \methodname.

\section{Discussion}
    
\begin{wrapfigure}[16]{r}{0.44 \textwidth}  %
\vspace{-1pt}
    \centering
    \includegraphics[width=1.0\linewidth]{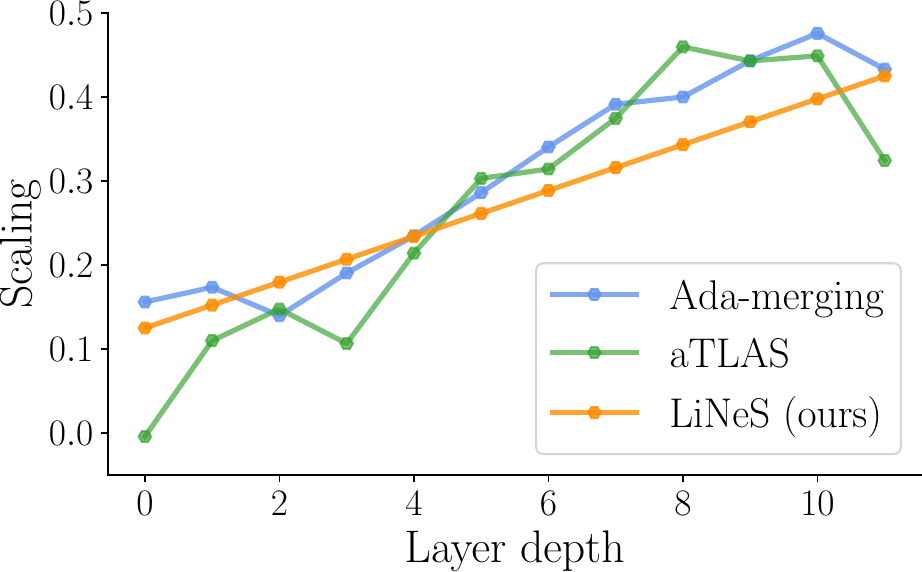}
    \caption{Comparison of the scalings obtained by different methods on 8-task merging benchmark with CLIP ViT-B/32. \looseness=-1}
    \label{fig:compare_scaling}
\end{wrapfigure}We compare \methodname with prior work that optimizes the scaling coefficients via backpropagation. Specifically, Ada-merging \citep{yang2023adamerging} minimizes the entropy loss of the predictions on the test set, while aTLAS \citep{zhang2024knowledge} 
minimizes a cross entropy loss on validation samples. Both methods operate on a more fine-grained level and introduce coefficients per layer \textit{and} per task, requiring all $T+1$ checkpoints, for task vectors and pre-trained model respectively, to be stored in memory during their fine-tuning process. %

We consider the 8-task computer vision benchmark with the ViT-B/32 visual encoder, and present the per-layer scalings in \autoref{fig:compare_scaling}. For aTLAS and Ada-merging, we report the average optimized scaling coefficients for attention and linear layers in each block across tasks. Without requiring training, \algoname leverages the inductive bias of neural networks to achieve scaling very close to Ada-merging or aTLAS , but with much less computational cost. Apart from the excessive memory overhead, both aTLAS and Ada-merging require multiple training epochs, making it challenging to scale for large models. As we demonstrate in \autoref{sec:rewarded_soup}, \algoname efficiently scales to large models like LLaMA \citep{Touvron_Martin_Stone_etal_2023}.

\section{Conclusion}

In this work, we presented \algoname, a novel method designed to mitigate catastrophic forgetting after fine-tuning process. By reducing the magnitude for parameter updates in the shallower layers, \algoname improves the generalization performance of the edited model on control tasks while almost fully preserving performance on the fine-tuned tasks. Furthermore, we demonstrated the versatility of \algoname in addressing task interference in multi-task model merging, where it consistently improves the baseline model merging methods across vision and NLP benchmarks. Our experiments confirm the broad applicability of \algoname across various scenarios, from improving OOD generalization to enhancing multi-task and single-task model merging strategies, as well as improving merging LLM policies aligned with different rewards. Given its simplicity and ease of integration with existing methods, \algoname offers a practical and inexpensive solution for boosting the generalization and robustness of fine-tuned models in diverse application domains.
\looseness=-1

\section*{Acknowledgments}
The authors thank Adam Hazimeh and the anonymous reviewers for their constructive discussions and comments.%

\clearpage

\clearpage

\bibliography{references}
\bibliographystyle{iclr2025_conference}

\clearpage
\appendix
\addcontentsline{toc}{section}{Appendix} %
\part{Appendix} %
\parttoc %

\clearpage
\section{\methodname Pseudocode}
\label{appendix:pseudocode}

We provide here a python pseudocode for the scaling the task vectors.

\begin{scriptsize} %
\begin{minted}[bgcolor=white]{python}
def line_scaling(task_vector, alpha=0.0, beta=1.0, num_blocks=12):
    """
    Progressively scales the task vector based on layer depth.

    Parameters:
    -----------
    task_vector : dict
        A dictionaryves control performantween the fine-tuned
        checkpoint and the pre-trained checkpoint.
    alpha : float
         The minimum scaling factor for the blocks.
    beta : float
        The maximum scaling coefficient difference between the last and first
        block.
    num_blocks : int
        The total number of layer blocks in the model.
    Returns:
    --------
    scaled_task_vector : dict
        A copy of `task_vector` where each key is scaled based on the layer
        depth.
    """

    import copy

    # Deep copy the task vector to avoid modifying the original
    scaled_task_vector = copy.deepcopy(task_vector)

    # Generate the key blocks corresponding to the layers of the model
    key_blocks = [f".layer{i}." for i in range(num_blocks)]

    # Create a scaling dictionary to store the scaling factor for each key
    scaling_dic = {}
    for k in task_vector.keys():
        # Find the layer block in the key and assign scaling factor based
        # on layer depth
        for layer, block in enumerate(key_blocks):
            if block in k:
                scaling_dic[k] = alpha + beta * (layer / (num_blocks - 1))
                break

    # Scale the task vector based on the scaling dictionary
    scaled_task_vector.vector = {
        # Use alpha if layer is outside residual blocks
        k: task_vector.vector[k] * scaling_dic.get(k, alpha)
        for k in task_vector.keys()
    }

    return scaled_task_vector


# example: scale single-task fine-tuned residual
task_vector = {k: theta_t[k] - theta_0[k] for k in theta_0.keys()}
scaled_task_vector = line_scaling(
    task_vector, alpha=gamma, beta=1.0 - gamma, num_blocks=12
)

# example: Scale the multi-task vectors
mtv = {
    k: sum(theta_ft[k] - theta_0[k] for theta_ft in ft_models)
    for k in theta_0.keys()
}
scaled_mtv = line_scaling(
    mtv, alpha=1 / len(ft_models), beta=beta, num_blocks=12
)
\end{minted}
\end{scriptsize} %

\section{Experimental Details}
\label{appendix:experimental details}

\subsection{Descriptions of baseline model merging methods}
\label{app:baselines}

\begin{itemize}
    \item \textbf{Task Arithmetic} \citep{ilharco2023task} generates a multi-task vector by summing the individual task vectors for each task. This multi-task vector is then added to the pre-trained checkpoint, with a scaling factor chosen based on validation set performance.
    \item \textbf{Ties-Merging} \citep{yadav2023ties} resolves parameter conflicts during model merging by first pruning parameters with lower magnitudes from the individual task vectors, followed by addressing sign mismatches, and finally merging parameters with consistent signs with averaging operation. The resulting multi-task vector is then added to the pre-trained checkpoint using a scaling factor determined from the validation set.
    \item \textbf{Consensus Merging} \citep{wang2024localizing} enhances existing model merging techniques by eliminating redundant weights in the multi-task vector. It first identifies the relevant subset of parameters for each task, then filters out weights that are relevant to either none or only one task. After removing these redundant weights, the refined multi-task vector is added to the pre-trained checkpoint with a scaling factor selected from the validation set. 
\end{itemize}

While consensus merging can be applied to various merging methods, in all our experiments, we evaluate only its application to task arithmetic.

\subsection{Experimental details for multi-task model merging}
\label{app:experimental_detail_multi_task}

\subsubsection{Hyper-parameters tuning}
\label{app:hyper_multi_task}

We list here the hyper-parameter search space for each model merging method in Table~\ref{tab:hyper_params}, while we suggest the authors to the original papers for a detailed description of these hyper-parameters. We highlight that applying \algoname does not introduce extra computational cost in hyper-parameter search for the baseline merging methods.

\begin{table}[ht!]

\resizebox{\textwidth}{!}{%

\begin{tabular}{@{}ccl@{}}
\toprule
\multirow{2}{*}{Method} & \multirow{2}{*}{\begin{tabular}[c]{@{}c@{}}With \\ \algoname\end{tabular}} & \multicolumn{1}{c}{\multirow{2}{*}{Hyper-parameter search space}} \\
 &  & \multicolumn{1}{c}{} \\ \midrule
\multirow{2}{*}{Task Arithmetic} & \ding{55} & constant scaling term for multi-task vector: [0.1,0.2,0.3,0.4,0.5,0.6,0.7,0.8,0.9,1.0] \\ \cmidrule(l){2-3} 
 & \ding{51} & scaling term $\beta$ in Eq.~\ref{eq:linear_scaling_a_b} for the multi-task vector: [0.1,0.2,0.3,0.4,0.5,0.6,0.7,0.8,0.9,1.0] \\ \midrule
\multirow{2}{*}{Ties-Merging} & \ding{55} & constant scaling term for multi-task vector: [0.1,0.2,0.3,0.4,0.5,0.6,0.7,0.8,0.9,1.0,1.1,1.2,1.3,1.4,1.5] \\ \cmidrule(l){2-3} 
 & \ding{51} & scaling term $\beta$ in Eq.~\ref{eq:linear_scaling_a_b} for the multi-task vector: [0.1,0.2,0.3,0.4,0.5,0.6,0.7,0.8,0.9,1.0,1.1,1.2,1.3,1.4,1.5] \\ \midrule
\multirow{2}{*}{Consensus Merging} & \ding{55} & \begin{tabular}[c]{@{}l@{}}constant scaling term for multi-task vector: [0.1,0.2,0.3,0.4,0.5,0.6,0.7,0.8,0.9,1.0];\\ weight-pruning threshold: [1, 2]\end{tabular} \\ \cmidrule(l){2-3} 
 & \ding{51} & \begin{tabular}[c]{@{}l@{}}scaling term $\beta$ in Eq.~\ref{eq:linear_scaling_a_b} for the multi-task vector: [0.1,0.2,0.3,0.4,0.5,0.6,0.7,0.8,0.9,1.0]; \\ weight-pruning threshold: [1, 2]\end{tabular} \\ \bottomrule
\end{tabular}
}
\end{table}
\label{tab:hyper_params}

\subsubsection{Benchmarks}
\label{app:benchmark contents}

\paragraph{Image classification}

For the benchmarks used in image classification, we utilized the 8-task benchmark initially proposed by \citet{ilharco2023task}, as well as the 14-task and 20-task benchmarks expanded by \citet{wang2024localizing}.

\begin{itemize}
    \item The \textbf{8-task benchmark} comprises the following tasks: Cars \citep{krause20133d}, DTD \citep{dtd}, EuroSAT \citep{eurosat}, GTSRB \citep{gtsrb}, MNIST \citep{lecun1998mnist}, RESISC45 \citep{cheng2017remote}, SUN397 \citep{sun397}, and SVHN \citep{svhn}.
    \item The \textbf{14-task benchmark} includes the original eight tasks plus additional ones: CIFAR100 \citep{krizhevsky2009learning}, STL10 \citep{stl10}, Flowers102 \citep{nilsback2008automated}, OxfordIIITPet \citep{parkhi2012cats}, PCAM \citep{veeling2018rotation}, and FER2013 \citep{goodfellow2013challenges}.
    \item The \textbf{20-task benchmark} builds on the 14-task benchmark with the addition of: EMNIST \citep{cohen2017emnist}, CIFAR10 \citep{krizhevsky2009learning}, Food101 \citep{food101}, FashionMNIST \citep{fashionmnist}, RenderedSST2 \citep{socher2013recursive,radford2019language}, and KMNIST \citep{clanuwat2018deep}.
\end{itemize}

\paragraph{Natural Language Processing}

For our NLP experiments, we utilized benchmarks established by \citet{yadav2023ties}, \citet{Tam_Bansal_Raffel_2023}, and \citet{wang2024localizing}.

\begin{itemize}
    \item The \textbf{7 NLP Tasks} benchmark, as explored in \citet{yadav2023ties}, includes the following datasets: QASC \citep{allenai:qasc}, QuaRTz \citep{tafjord-etal-2019-quartz}, PAWS \citep{zhang-etal-2019-paws}, Story Cloze \citep{sharma-etal-2018-tackling}, WikiQA \citep{yang-etal-2015-wikiqa}, Winogrande \citep{WinoGrande}, and WSC \citep{levesque2012winograd}.
    \item The \textbf{8 QA Tasks} \citep{Tam_Bansal_Raffel_2023} comprises the following datasets: CosmosQA \cite{huang2019cosmos}, QASC \citep{allenai:qasc}, QuAIL \cite{rogers2020getting}, QuaRTz \citep{tafjord-etal-2019-quartz}, PAWS \citep{zhang-etal-2019-paws}, ROPES \cite{lin2019reasoning}, SocialIQA \cite{sap2019socialiqa}, and WikiQA \citep{yang-etal-2015-wikiqa}.
    \item The \textbf{11 NLP Tasks} benchmark is a union of these two benchmarks, as studied in \cite{wang2024localizing}. It contains the following tasks: QASC \citep{allenai:qasc}, QuaRTz \citep{tafjord-etal-2019-quartz}, PAWS \citep{zhang-etal-2019-paws}, Story Cloze \citep{sharma-etal-2018-tackling}, WikiQA \citep{yang-etal-2015-wikiqa}, Winogrande \citep{WinoGrande}, WSC \citep{levesque2012winograd}, CosmosQA \cite{huang2019cosmos}, QuAIL \cite{rogers2020getting}, ROPES \cite{lin2019reasoning}, and SocialIQA \cite{sap2019socialiqa}. 
\end{itemize}

\subsection{Experimental details for single-task model merging}
\label{app:experimental_details_single_task}

\subsubsection{Description of Model Soups}

\textbf{Model soups} \citep{wortsman2022model} is a model merging method which averages the weights of multiple fine-tuned models with different hyper-parameter configurations, improving accuracy of the merged model without increasing inference or memory costs. The authors of model soups proposed two methods:

\begin{itemize}
    \item \textbf{Uniform soup}: Averages the weights of all fine-tuned checkpoints, providing a simple and efficient way to improve performance.
    \item \textbf{Greedy soup}: Starting with the best-performing checkpoint, greedily and iteratively adds the next best-performing checkpoint to the soup, keeping those that improve accuracy of current collection of model checkpoints.
\end{itemize}

\subsubsection{Experimental details for modifications to model soups}

We describe in detail the modifications to model soups, namely task arithmetic and our proposed \algoname. For reference, model soups merges the checkpoints by averaging the weights of the individual checkpoints:

\begin{equation}
    \bm{\theta}_{\text{soup}}^{\text{vanilla}}
    =\bm{\theta}_0 + \bm{\tau}_{\text{soup}}
\label{eq:app_soup}
\end{equation}

\paragraph{Enhancing Model Soups with Task Arithmetic} 
We enhance model soups with task arithmetic, by introducing a scaling factor $\lambda_{\text{ta}}$ to $\bm{\tau}_{\text{soup}}$ in Equation~\ref{eq:app_soup}. We search for this hyper-parameter within the range of $[1.0, 1.1, 1.2, 1.3, 1.4, 1.5, 1.6, 1.7, 1.8, 1.9, 2.0]$. Note that $\lambda_{\text{ta}} = 1.0$ yields the vanilla model soups.

\begin{equation}
    \bm{\theta}_{\text{soup}}^{\text{ta}}
    =\bm{\theta}_0 + \lambda_{\text{ta}} \cdot \bm{\tau}_{\text{soup}}
\end{equation}

\paragraph{Enhancing Model Soups with \algoname} 

We enhance model soups with \algoname, by applying \algoname to $\bm{\tau}_{\text{soup}}$ in Equation~\ref{eq:app_soup}. For the scaling, we apply directly the scaling introduced in Equation~\ref{eq:linear_scaling_a_b} to create a scaled task vector $\bm{\tau}_{\text{soup}}^{\text{LiNeS}}$, fixing $\alpha$ to $1$ while searching the value for $\beta$ within the range of [0.0, 0.1, 0.2, 0.3, 0.4, 0.5, 0.6, 0.7, 0.8, 0.9, 1.0]. Note that $\beta = 0.0$ yields the vanilla model soups. 

\begin{equation}
    \bm{\theta}_{\text{soup}}^{\text{LiNeS}}
    =\bm{\theta}_0 + \bm{\tau}_{\text{soup}}^{\text{LiNeS}}
\end{equation}

We further note that, both Task Arithmetic and our proposed method introduce only one hyper-parameter to model soups, while the computational cost for hyper-parameter search is the same. When applying the modifications to greedy soup, we only apply them directly to the selected subset of checkpoints after the greedy selection process.

We search for the hyper-parameter within the validation set and report the performance on test set with the best hyper-parameter based on validation performance.

\section{Additional Results}
\label{app:addition_result}

\subsection{Different trade-offs for retention of task and control task performance}
\label{app:trade_off_selection}

In Section~\ref{sec:motivation}, we need to balance two competing objectives: maximizing accuracy on the target task while preserving performance on the control task. To account for different user preferences, we scalarize these objectives by assigning varying weights to the target task accuracy. This weighting scheme can be adjusted depending on the scenario to reflect different priorities. Let $w_\text{target}$ represent the weight assigned to the target task accuracy, and $\text{M}_\text{target}$ and $\text{M}_\text{control}$ denote the normalized accuracies for the target and control tasks, respectively. The optimal value of $\gamma$ is selected to maximize the following weighted trade-off on the validation set: %
\[
w_\text{target}\text{M}_\text{target} + \text{M}_\text{control}
\]
To account for the high variance in control task performance and to emphasize the target task, we assign it a weight of 2, signifying that its accuracy is prioritized twice as much as the control task's accuracy.
\looseness=-1

\begin{table}[h]
\caption{Validation results on the target vs control performance benchmark, presented in \autoref{sec:motivation}, averaged over the 8 tasks. We balance two competing objectives with various scalarization weights $w_\text{target}.$ In the main text, we use $w_\text{target}=2$.}
\label{tab:importance_weight}
\centering
\begin{tabular}{@{}ccc@{}}
\toprule
\multirow{2}{*}{$w_\text{target}$} & \multicolumn{2}{c}{Averaged normalized accuracy (\%)} \\ \cmidrule(l){2-3} 
 & Target task & Control tasks \\ \midrule
1 & 99.8 & 101.9 \\
2 & 100.0 & 101.5 \\
5 & 100.2 & 100.8 \\ \bottomrule
\end{tabular}

\end{table}

\subsection{Detailed labels for Figure~\ref{fig:id_ood_norm_acc}}

\begin{figure}[ht!]
    \centering
    \includegraphics[width=1\linewidth]{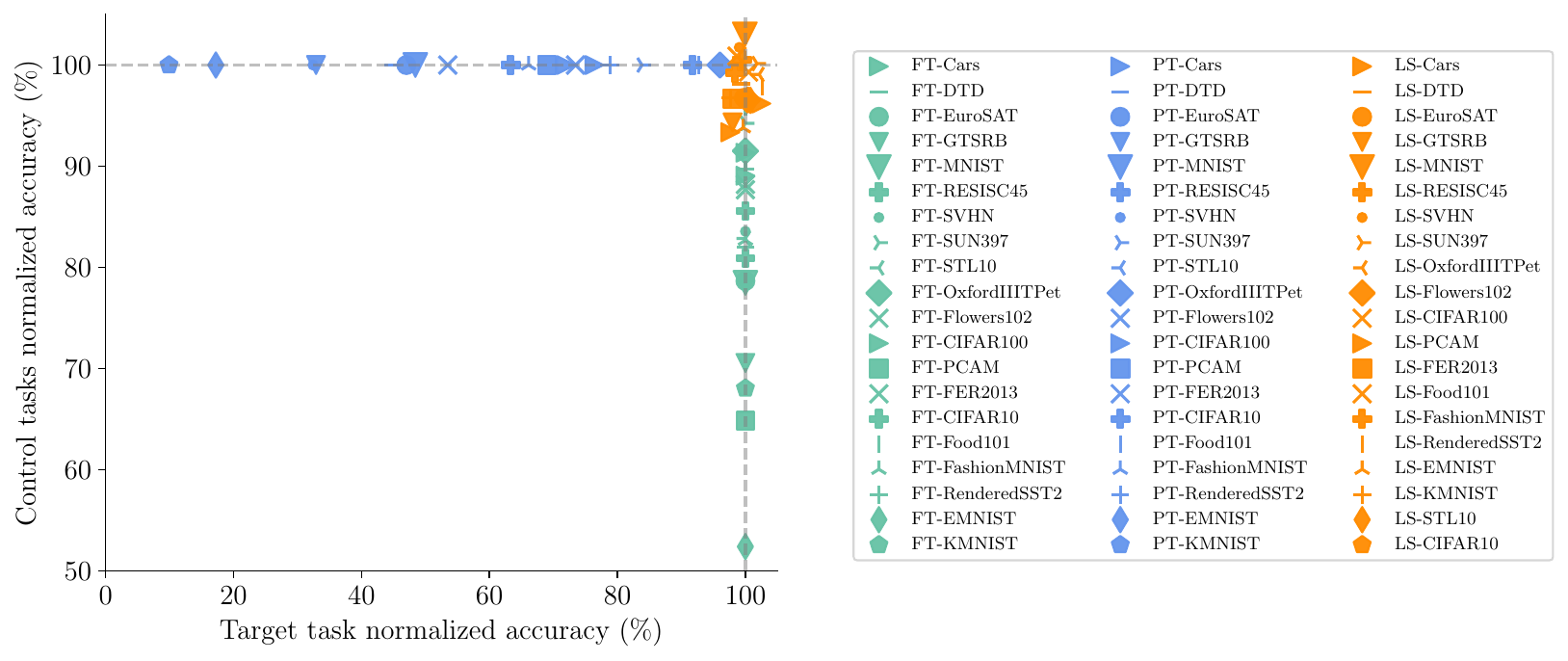}
    \caption{Figure~\ref{fig:id_ood_norm_acc} with detailed label information. Each scatter dot corresponds to applying a specific model (FT for fine-tuned model; PT for pre-trained model; LS for fine-tuned model edited with \algoname) on different task.}
    \label{fig:id_ood_norm_acc_app}
\end{figure}

We provide in Figure~\ref{fig:id_ood_norm_acc_app} the detailed labels corresponding to each scatter dot. Each scatter dot corresponds to applying a specific model (FT for fine-tuned model; PT for pre-trained model; LS for fine-tuned model edited with \algoname) on different tasks.

\subsection{Ablations of different choices of scaling function}
\label{app:ablation_scaling}

We provide in this section an ablation study on applying different scaling functions for \algoname. In \algoname we used directly  $\lambda^{(\ell)} = \alpha + \beta \cdot \frac{\ell-1}{\numlayers-1}$ to scale different layers. Here we test the performance on multi-task model merging in vision benchmarks with the following choices for scaling functions $f(\cdot)$: linear scaling, quadratic scaling and square root scaling:

\begin{itemize}
    \item linear scaling: $\lambda^{(\ell)} = \alpha + \beta \cdot \frac{\ell-1}{\numlayers-1}$
    \item square root scaling: $\lambda^{(\ell)} = \alpha + \beta \cdot {\left( \frac{\ell-1}{\numlayers-1} \right)}^{\frac{1}{2}}$ 
    \item quadratic scaling: $\lambda^{(\ell)} = \alpha + \beta \cdot {\left( \frac{\ell-1}{\numlayers-1} \right)}^{2}$
\end{itemize}

We provide in Table ~\ref{tab:ablation_scaling} the performance of different choices of scaling on vision benchmarks with ViT-B/32. While using quadratic scaling sometimes outperforms using identify function, especially with a larger number of tasks during merging, the improvement is not substantial. Therefore, we choose the linear scaling to keep the method simple and general.

\begin{table}[ht!]
\centering
\caption{Ablation study for applying different scalings for \algoname on vision benchmarks with ViT-B/32.}
\resizebox{0.7\textwidth}{!}{%
\begin{tabular}{@{}ccccccc@{}}
\toprule
\multirow{2}{*}{Method} & \multirow{2}{*}{Scaling function} &  &  & \multicolumn{3}{c}{ViT-B/32} \\ \cmidrule(l){5-7} 
 &  &  &  & 8 tasks & 14 tasks & 20 tasks \\ \midrule
\multirow{3}{*}{Task Arithmetic} & linear &  &  & \textbf{74.2} & 69.1 & 63.4 \\
 & square root &  &  & 73.9 & 67.5 & 62.4 \\
 & quadratic &  &  & 73.8 & \textbf{69.2} & \textbf{64.6} \\ \midrule
\multirow{3}{*}{Ties-Merging} & linear &  &  & \textbf{77.2} & \textbf{72.1} & 67.2 \\
 & square root &  &  & 76.9 & 70.4 & 65.6 \\
 & quadratic &  &  & 76.1 & 71.6 & \textbf{67.4} \\ \midrule
\multirow{3}{*}{Consensus Merging} & linear &  &  & \textbf{77.6} & 73.6 & 68.6 \\
 & square root &  &  & 77.1 & 72.5 & 67.3 \\
 & quadratic &  &  & 77.1 & \textbf{73.9} & \textbf{69.0} \\ \bottomrule
\end{tabular}
}
\end{table}
\label{tab:ablation_scaling}

\subsection{Results for ViT-B/16 for multi-task merging}
\label{app:vit_b_16}

We provide in Table~\ref{tab:vision_app} the results complementary to Table~\ref{tab:vision} for using ViT-B/16 as the image encoder, where we observe similar performance gains and observations by using \algoname as in Table~\ref{tab:vision}.

\begin{table}[ht]
\centering
\caption{Complementary to Table~\ref{tab:vision}, for results obtained with ViT-B/16 as image encoder.}
\resizebox{0.72\textwidth}{!}{%
\begin{tabular}{@{}lccclll@{}}
\toprule
\multirow{2}{*}{Method} & \multirow{2}{*}{\begin{tabular}[c]{@{}c@{}}with \\ \algoname\end{tabular}} &  &  & \multicolumn{3}{c}{ViT-B/16} \\ \cmidrule(l){5-7} 
 &  &  &  & 8 tasks & 14 tasks & 20 tasks \\ \midrule
\rowcolor[HTML]{EFEFEF} Zero-shot &  &  &  & 55.5 & 61.4 & 59.8 \\
\rowcolor[HTML]{EFEFEF} Fine-tuned &  &  &  & 92.6 & 91.6 & 92.3 \\ \midrule
\multirow{2}{*}{Task Arithmetic} & \ding{55} &  &  & 74.6 & 70.4 & 65.7 \\
 & \ding{51} &  &  & \textbf{\foo{77.6}{3.0}} & \textbf{\foo{72.7}{2.3}} & \textbf{\foo{67.7}{2.0}} \\ \midrule
\multirow{2}{*}{Ties-Merging} & \ding{55} &  &  & 79.1 & 73 & 68.1 \\
 & \ding{51} &  &  & \textbf{\foo{79.9}{0.8}} & \textbf{\foo{75.2}{2.2}} & \textbf{\foo{71.2}{3.1}} \\ \midrule
\multirow{2}{*}{Consensus Merging} & \ding{55} &  &  & 78.9 & 73.9 & 70.2 \\
 & \ding{51} &  &  & \textbf{\foo{79.5}{0.6}} & \textbf{\foo{75.8}{1.9}} & \textbf{\foo{72.0}{1.8}} \\ \bottomrule
\end{tabular}
}

\end{table}
\label{tab:vision_app}

\subsection{Results for editing T5 with \algoname}
\label{app:shallow_t5}

We repeat here similar experiments we performed on ViT-B/32 model in Section~\ref{sec:motivation} with T5-large \citep{colin2020exploring}. T5-large contains both encoder and decoder structure, with sequential residual blocks in both structures. We investigate separately how the shallow-layer updates in the encoder and decoder infect the target and control task accuracy.

\begin{figure}[ht!]
    \centering
    \includegraphics[width=1\linewidth]{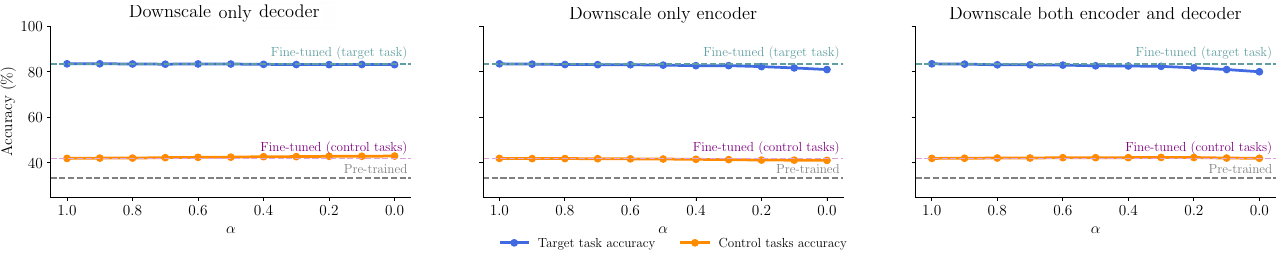}
    \caption{The impact of downscaling the shallower-layer parameter updates on T5-large model within the 8-question-answering benchmark \citep{colin2020exploring}. Downscaling only on the decoder (left) architecture preserves full target performance, while slightly improving the control tasks performance. Downscaling on the encoder leads to performance degradation on target tasks.}
    \label{fig:redundancy_t5}
\end{figure}

We consider the 8-question-answering benchmark \citep{zhou2022not}, and plot in Figure~\ref{fig:redundancy_t5} the averaged target and control task accuracy after applying \algoname to 
\begin{enumerate}
    \item only the decoder part (left),
    \item only the encoder part (middle),
    \item both the encoder and decoder part (right).
\end{enumerate} 

We observe that, only downscaling on the decoder architecture fully preserves full target performance, while downscaling on the encoder, or on both encoder and decoder, leads to performance drop on the target tasks. On the other hand, downscaling on the decoder part slight improves control generalization of the fine-tuned model, which we do not observe from downscaling on the encoder, or simultaneously on the encoder and decoder. We also note that, unlike the case in vision, the fine-tuned checkpoints on this NLP benchmark actually improve over the zero-shot performance of the pre-trained model on control tasks.

These results motivate us to apply \algoname to only the decoder part of T5-large when merging multiple checkpoints, which preserves full target task accuracy while slightly improving control task performance, leading to similar observation in applying \algoname to the ViT-B/32 architecture in vision,

\subsection{Sensitivity analysis for hyper-parameters}
\label{app:sensitivity_analysis}

We provide in this section the sensitivity analysis for the hyper-parameters of \algoname. Specifically, we consider the setting in multi-task merging in the 8-task vision classification benchmark with ViT-B/32 CLIP model. 

\begin{figure}[h]
    \centering
    \includegraphics[width=0.9\linewidth]{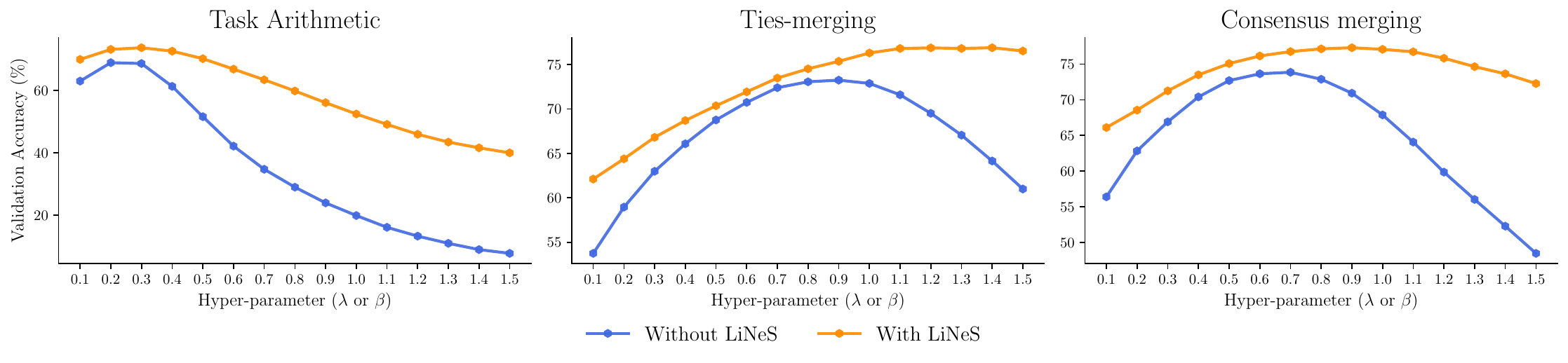}
    \caption{Sensitivity to hyper-parameters in multi-task merging for the 8-task benchmark with CLIP ViT-B/32 model. The y-axis represents the averaged multi-task validation accuracy and x-axis represents the hyper-parameter value, i.e., $\lambda$ for the baseline method and $\beta$ for the method enhanced with \algoname. \looseness=-1}
    \label{fig:sensitivity_lambda_vs_beta}
\end{figure}

As explained in \autoref{sec:multi-task model merging}, \algoname fixes $\alpha$ with a heuristic value by \autoref{eq:heuristic} and only tunes $\beta$ for multi-task merging. The slope hyper-parameter $\beta$ is tuned within the same range as the uniform scaling coefficient $\lambda$ for the baseline merging methods. We compare in \autoref{fig:sensitivity_lambda_vs_beta} the sensitivity of averaged multi-task validation accuracy to the respective hyper-parameters, i.e., to $\lambda$ for baseline merging methods and $\beta$ for the \algoname-enhanced merging methods. The results show that, for all three merging methods, including Task arithmetic, Ties-merging and Consensus, enhancing with \algoname is less sensitive to hyper-parameter choices compared to the corresponding baseline method. 

\begin{figure}[h]
    \centering
    \includegraphics[width=\linewidth]{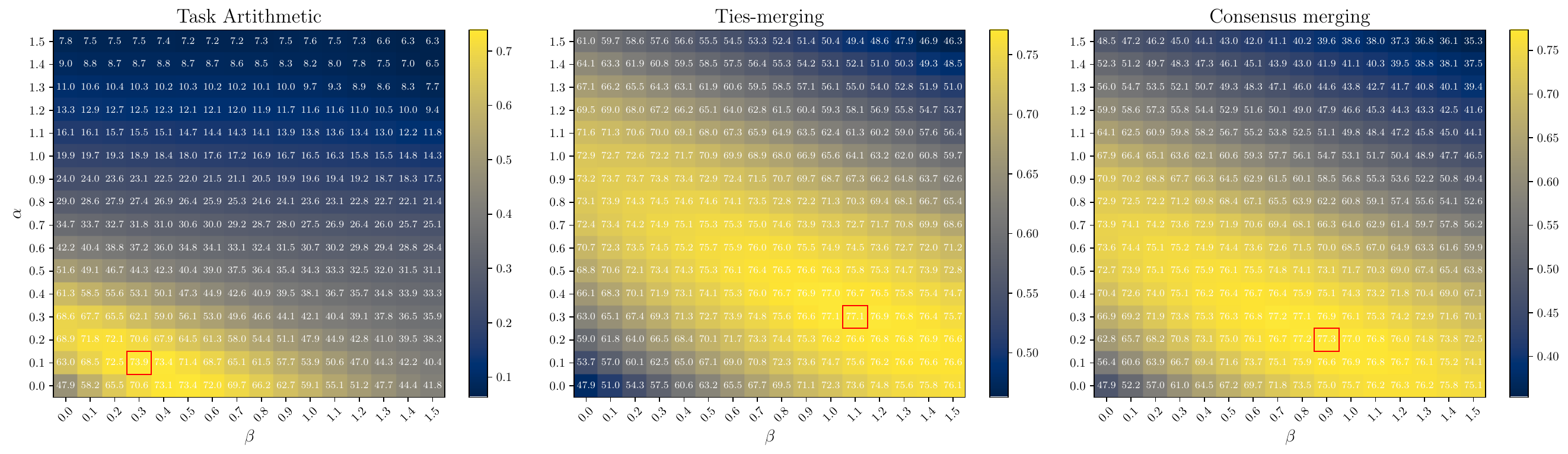}
    \caption{Sensitivity to both $\alpha$ and $\beta$ in multi-task merging for the 8-task benchmark with CLIP ViT-B/32 model. The heatmap represents the averaged multi-task validation accuracy, while x and y axis represent the $\beta$ and $\alpha$ respectively.  The optimal configuration is annotated with a red box.}
    \label{fig:sensitivity}
\end{figure}

Furthermore, we perform an ablation treating $\alpha$ as a hyper-parameter and analyze the sensitivity to both $\alpha$ and $\beta$ for \algoname in a two-dimensional grid for the same benchmark. The results are presented in \autoref{fig:sensitivity}. The results clearly demonstrate the necessity for applying layer-increasing scaling, as the optimal performance is obtained with both $\alpha > 0$ and $\beta > 0$ for all three merging method. 
Note that the optimal configurations found by the ablation study are very close to the configurations found in our method, as shown in \autoref{tab:heuristic}, by setting $\alpha$ via the heuristic and searching only for $\beta$. 
\looseness=-1

\begin{table}[h]
    \centering
    \caption{$\alpha$ and $\beta$ values for $\alpha$ set by our proposed heuristic in \autoref{eq:heuristic}.}
    \label{tab:heuristic}
    \begin{tabular}{c|cc}
    \toprule
     Method         & $\alpha$ & $\beta$ \\
     \midrule
     Task Arithmetic    & 0.125 & 0.3 \\
     Ties-merging       & 0.21 & 1.4 \\
     Consensus merging  & 0.21 & 0.9 \\
     \bottomrule
    \end{tabular}
\end{table}

\subsection{Experiments with CNN architectures}

\begin{figure}[h]
    \centering
    \includegraphics[width=0.55\linewidth]{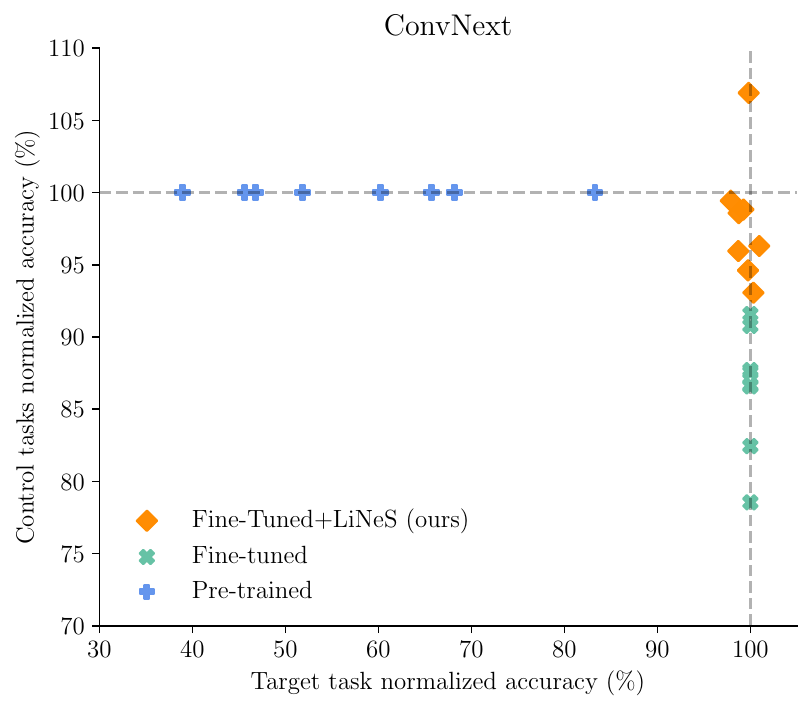}
    \caption{Our linear scaling (\algoname) retains performance on both control and fine-tuned target tasks for ConvNext architecture.}
    \label{fig:convnext}
\end{figure}

In this section, we apply \algoname to CNN architectures. Specifically, we consider the ConvNeXt \citep{liu2022convnet} architecture. First, we repeat the experiments presented in \autoref{sec:motivation} regarding mitigating catastrophic forgetting. The final results are presented in \autoref{fig:convnext}, where we observe similar findings with CLIP ViTs, i.e., \algoname greatly improves the performance on control tasks when applied to the fine-tuned checkpoints while preserving most of the accuracy on target tasks. 
Furthermore, we present in \autoref{tab:convnext-task arithmetic} the results on multi-task model merging, following the experimental protocol established in \autoref{sec:multi-task model merging}. Again, we see that \algoname improves the performance of baseline merging methods.

\begin{table}[h]
    \centering
        \caption{Multi-task model merging results using a ConvNeXt architecture. \algoname improves the results compared to uniform scaling for both Task Arithmetic and Ties-merging.}
    \label{tab:convnext-task arithmetic}
   \begin{tabular}{lclll}
\toprule
Method & LiNeS  & Acc (\%) & Norm. Acc (\%) &  \\
\midrule
\multirow[c]{2}{*}{Task Arithmetic} & \ding{55} & 77.9 & 83.8 &   \\
 & \ding{51} & 79.0 [+\textcolor{darkgreen}{1.1}] & 84.8 [+\textcolor{darkgreen}{1.0}] &   \\
\midrule
\multirow[t]{2}{*}{Ties-merging} & \ding{55} & 79.7 & 85.8 &   \\
 & \ding{51} & 80.3 [+\textcolor{darkgreen}{0.6}] & 86.3 [+\textcolor{darkgreen}{0.5}] &   \\
\bottomrule
\end{tabular}

\end{table}

\subsection{Experiments with Regularized Fine-Tuning}
\label{app:regularized_finetuning}

In this section, we evaluate \algoname against several regularized fine-tuning methods, focusing on their ability to preserve general features and mitigate catastrophic forgetting. The regularization strategies applied during fine-tuning are described below:

\begin{enumerate}
    \item \textbf{Fine-tuning with Linear Layer-Wise Learning Rate Decay (LinLR):} Applies a linear learning rate schedule where the learning rate linearly increases from 0.0 to the maximum value for all the layers.
    \item \textbf{Fine-tuning with Exponential Layer-Wise Learning Rate Decay (ExpLR):} Applies an exponential learning rate schedule where the learning rate is set to maximum for the deepest layers and decays by a factor of 0.5 by each layer for the shallower layers.
    \item \textbf{Fine-tuning with First Half of Blocks Frozen (HalfFT):} Freezes the parameters of the first half of the model’s blocks during training.
    \item \textbf{Fine-tuning only the Final Block (LastFT):} Freezes all blocks except the final block of the feature encoder.
\end{enumerate}

\begin{table}[h]
\caption{Performance of different methods on target and control tasks, averaged over all target and control task combinations in the 8-task vision benchmark \citep{ilharco2023task}.}
\resizebox{\linewidth}{!}{%
\begin{tabular}{@{}cccccccc@{}}
\toprule
\multicolumn{1}{l}{} & pre-trained & fine-tuned & FT+LiNeS & FT+LinLR & FT+ExpLR & FT+HalfFT & FT+LastFT \\ \midrule
Target (\%) & 48.3 & 90.5 & 90.3 & 90.7 & 89.6 & 90.4 & 85.6 \\
Control (\%) & 48.3 & 38.0 & 48.0 & 46.9 & 46.0 & 46.8 & 46.6 \\ \bottomrule
\label{table:regularized_ft}
\end{tabular}
}
\end{table}

We present the results in \autoref{table:regularized_ft}. We observe that \algoname, as a post-training editing method, outperforms the regularized fine-tuning methods in terms of restoring the zero-shot performance on the control tasks. We further emphasize that compared with the regularized fine-tuning methods, \algoname benefits from many advantages such as efficiency, flexibility and computational cost.

\subsection{Experiments with LoRA Fine-tuning}
\label{app:lora_finetuning}

We explore the applicability of the method on models fine-tuned with LoRA \citep{hu2022lora}. For a layer with pre-trained weights $\WW_0\in\r^{m\times n}$, LoRA adds trainable matrices $\AA\in\r^{m\times r}$ and $\BB\in\r^{n}$, for rank $r\lll \min(n, m)$. The weights of the layer become:
\begin{equation*}
    \WW = \WW_0 + \frac{\alpha}{r}\BB\AA
\end{equation*}
for $\alpha\in\r$. Following common practice, we set $\alpha=r$ and fine-tune with the same protocol used for full fine-tuning. We consider only the case of ViT-B/32 fine-tuned on 8 tasks and replicate the experiment presented in \autoref{tab:fine-tuning}.  Specifically, for each of the 8 LoRA-fine-tuned models, we compute the accuracy on the same (target) task as well as the average performance for each of the 7 remaining control tasks. \autoref{tab:lora ft} reports the average over the 8 cases for ranks $r\in\{16,32,64,128\}$. We observe that LoRA fine-tuning has lower target performance compared to full fine-tuning and that increasing target performance comes at the cost of more forgetting. In all the cases, \algoname restores control performance while minimally affecting target performance.

\begin{table}[h]
    \centering
        \caption{Similar to \autoref{tab:fine-tuning}, LoRA Fine-tuning harms generalization on control tasks. Increased target performance rsults in higher levels of forgetting. Still, our proposed method \algoname restores control performance for all ranks considered while minimally affecting target performance.}
    \label{tab:lora ft}
\begin{tabular}{lcc}
\toprule
 & Target & Control \\
\midrule
Pre-trained  & 48.3 & 48.3 \\
\midrule
Fine-tuned  & 90.5 & 38.0 \\ 
Fine-tuned $+ \texttt{LiNeS}$  & 90.3 & 48.0 \\ 
\midrule
$r=16$ & 84.4 & 44.2 \\
$r=16 + \texttt{LiNeS}$ & 84.3 & 46.7 \\
$r=32$ & 85.8 & 42.8 \\
$r=32 + \texttt{LiNeS}$ & 85.5 & 46.7 \\
$r=64$ & 86.6 & 41.6 \\
$r=64 + \texttt{LiNeS}$ & 86.4 & 46.2 \\
$r=128$ & 87.5 & 41.6 \\
$r=128 + \texttt{LiNeS}$ & 87.2 & 46.3 \\
\bottomrule
\end{tabular}

\end{table}

\subsection{Additional results for improving \wiseft with \algoname}
\label{app:robust_fine-tuning}

\subsubsection{Results for using ViT-B/16 as visual encoder}
\label{app:wiseft_vit_b_16}

\begin{figure}[ht]
    \centering
    \includegraphics[width=1.0\linewidth]{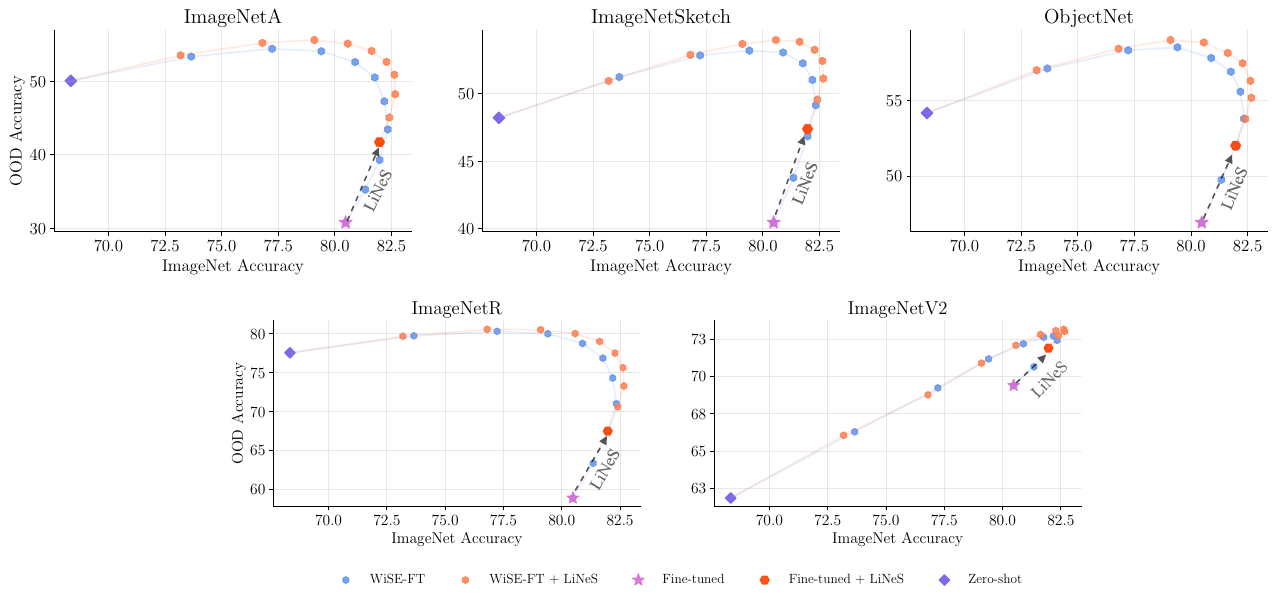}
    \caption{Results for improving \wiseft with \algoname on with ViT-B/16 model fine-tuned on ImageNet.}
    \label{fig:wiseft_vit_b_16}
\end{figure}

We provide in Figure~\ref{fig:wiseft_vit_b_16} the results for applying \algoname for improving \wiseft, using ViT-B/16 as visual encoder. The ViT-B/16 checkpoint obtained through fine-tuning the CLIP checkpoint on ImageNet with the same hyper-parameter configurations in \cite{wortsman2022robust}. From Figure~\ref{fig:wiseft_vit_b_16} we observe that \algoname improves over \wiseft for both ID and OOD accuracies, leading to similar observations as the results obtained with ViT-B/32.

\subsubsection{Individual results for 70 checkpoints}
\label{app:wiseft_individual}

\begin{figure}[ht]
    \centering
    \includegraphics[width=1\linewidth]{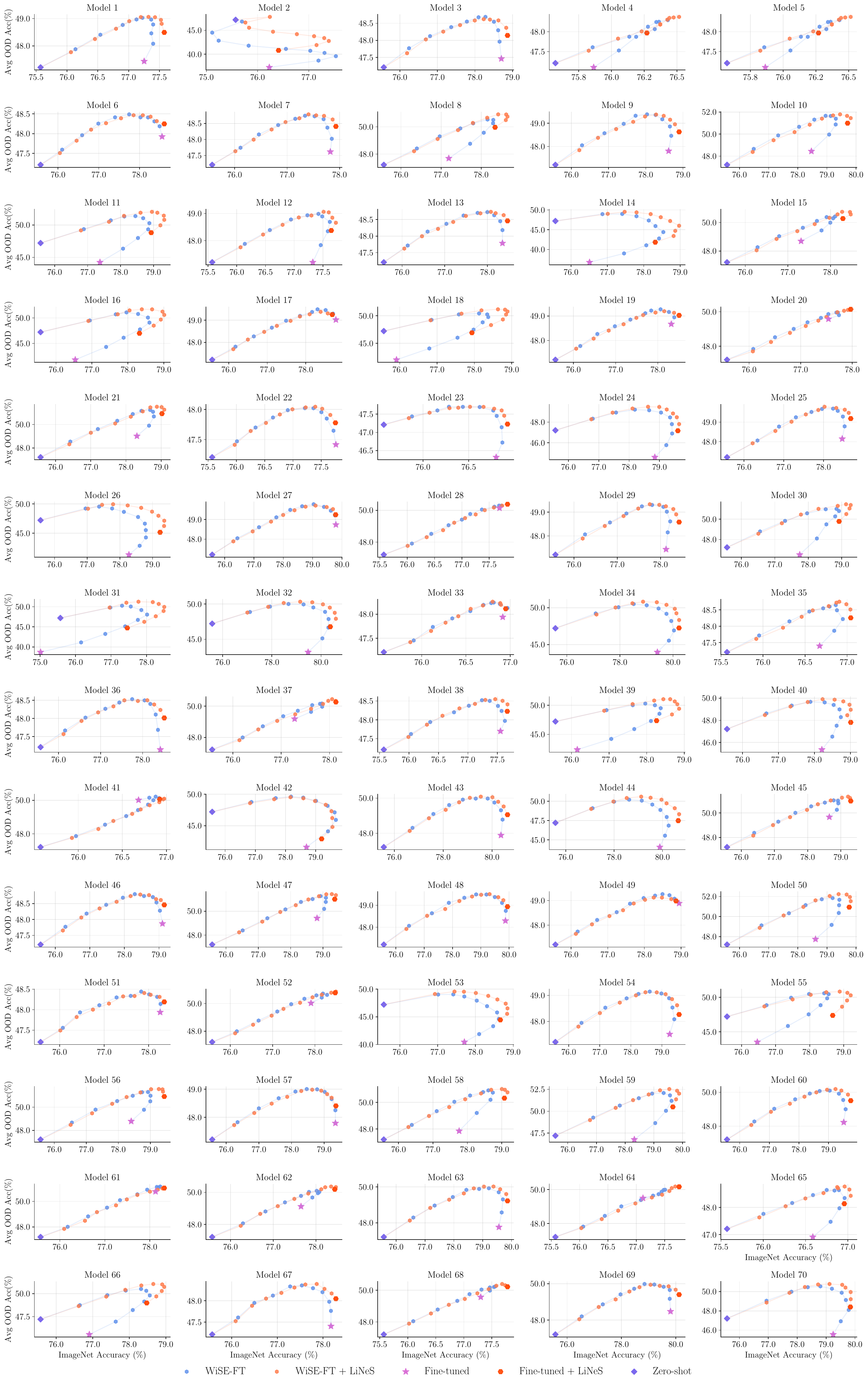}
    \caption{Performance of applying \algoname to \wiseft to each ViT-B/32 checkpoint fine-tuned on ImageNet.}
    \label{fig:robust_fine-tuning_individual}
\end{figure}

We provide in Figure~\ref{fig:robust_fine-tuning_individual} individual results separately for the experiments on the 70 individual model checkpoints. Note that here y-axis represents the averaged accuracy over 5 OOD datasets. From the figure, we observe that \algoname consistently improves \wiseft in terms of both ID and OOD accuracies for most of the individual checkpoints.

\subsection{Detailed performance on individual tasks for multi-task model merging}
\label{app:individual_performance}

\paragraph{Image Classification}

We provide the detailed performance on each individual task for multi-task model merging in image classification benchmarks, complementary to the results in Table~\ref{tab:vision} and Table~\ref{tab:vision_app} where the accuracies are averaged on the individual tasks. 

\begin{figure}[t]
    \centering
    \includegraphics[width=1\linewidth]{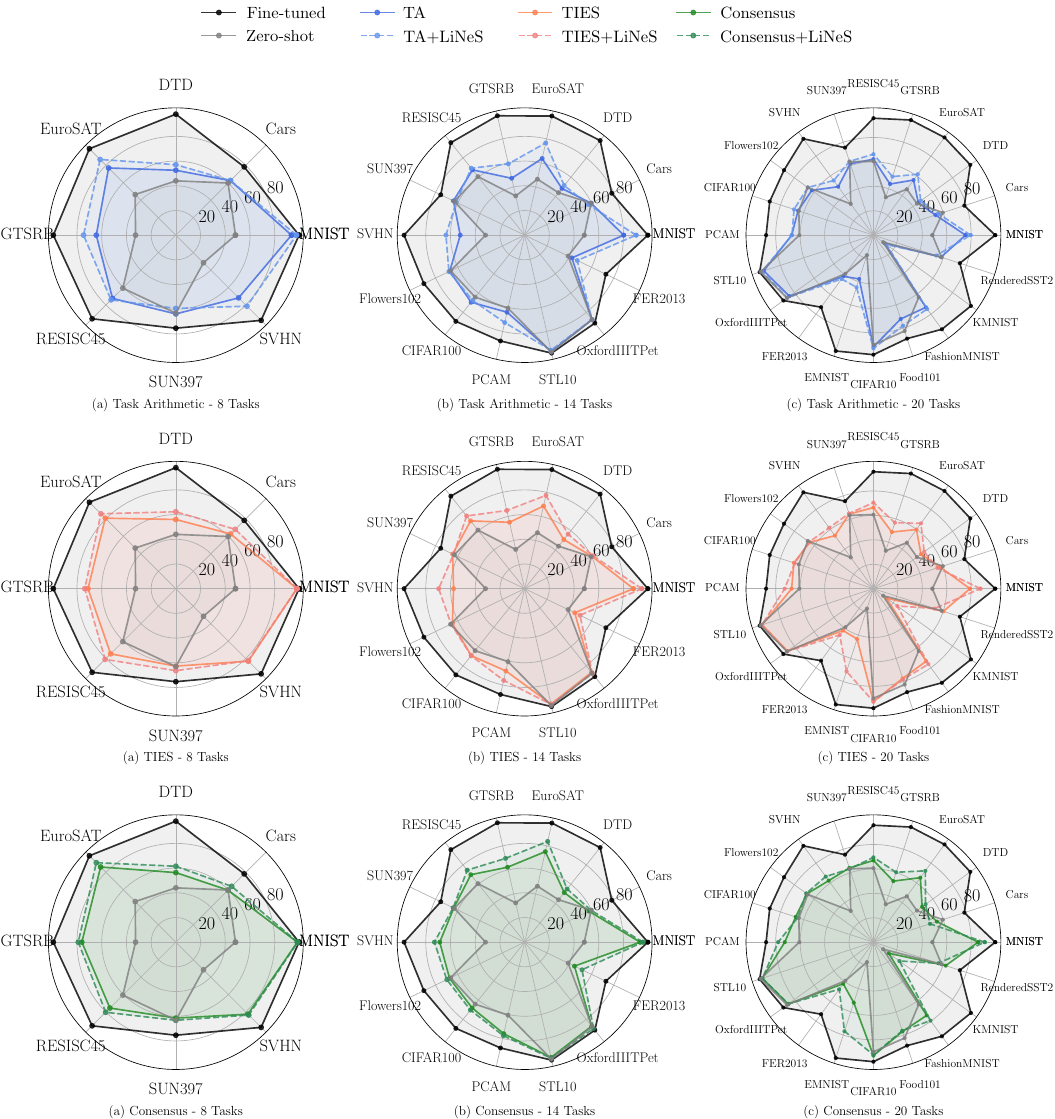}
    \caption{Single-task accuracies for multi-task merging on image classification benchmarks for ViT-B/32.}
    \label{fig:ind_vit_b_32}
\end{figure}

\begin{figure}[t]
    \centering
    \includegraphics[width=1\linewidth]{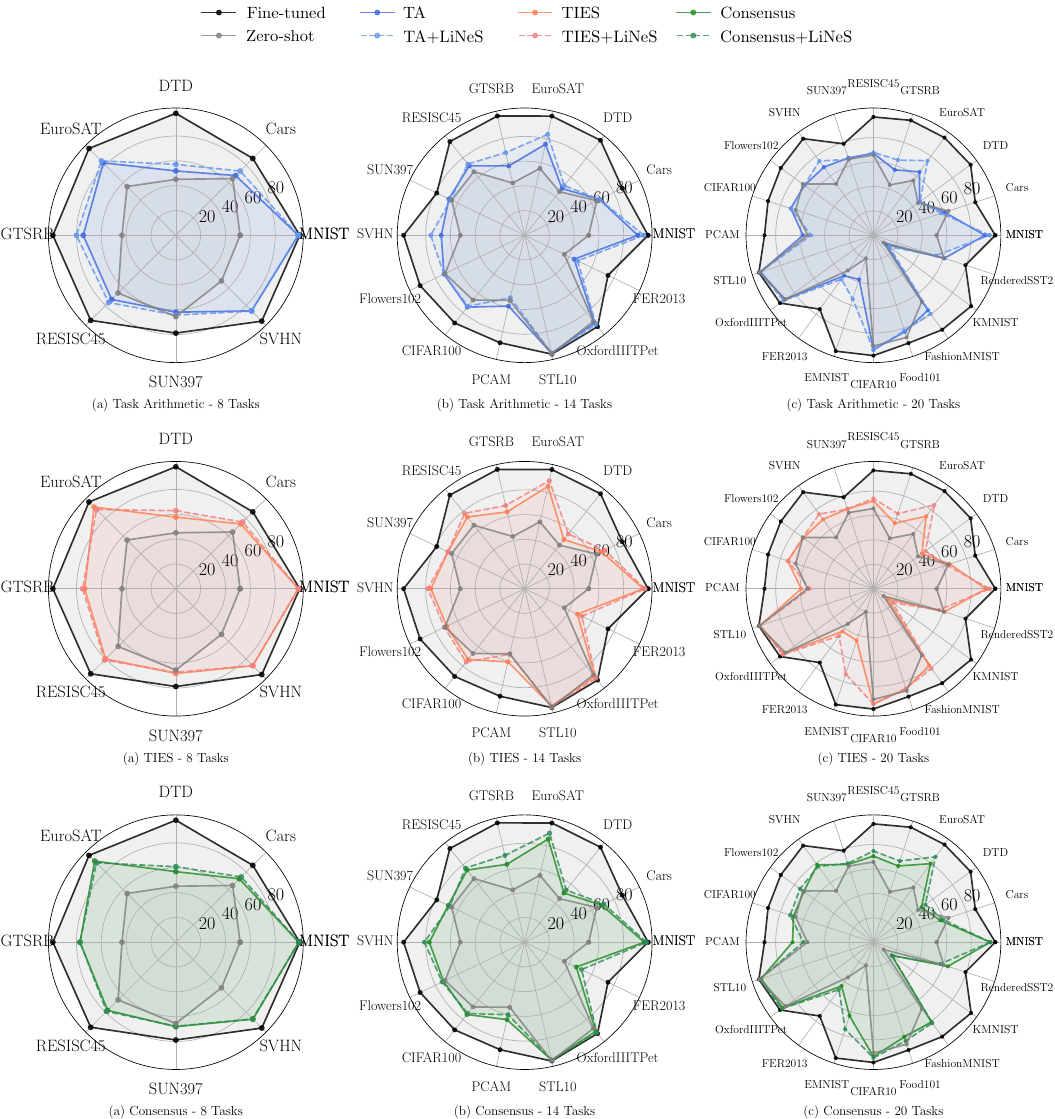}
    \caption{Single-task accuracies for multi-task merging on image classification benchmarks for ViT-B/16.}
    \label{fig:ind_vit_b_16}
\end{figure}

\begin{figure}[t]
    \centering
    \includegraphics[width=1\linewidth]{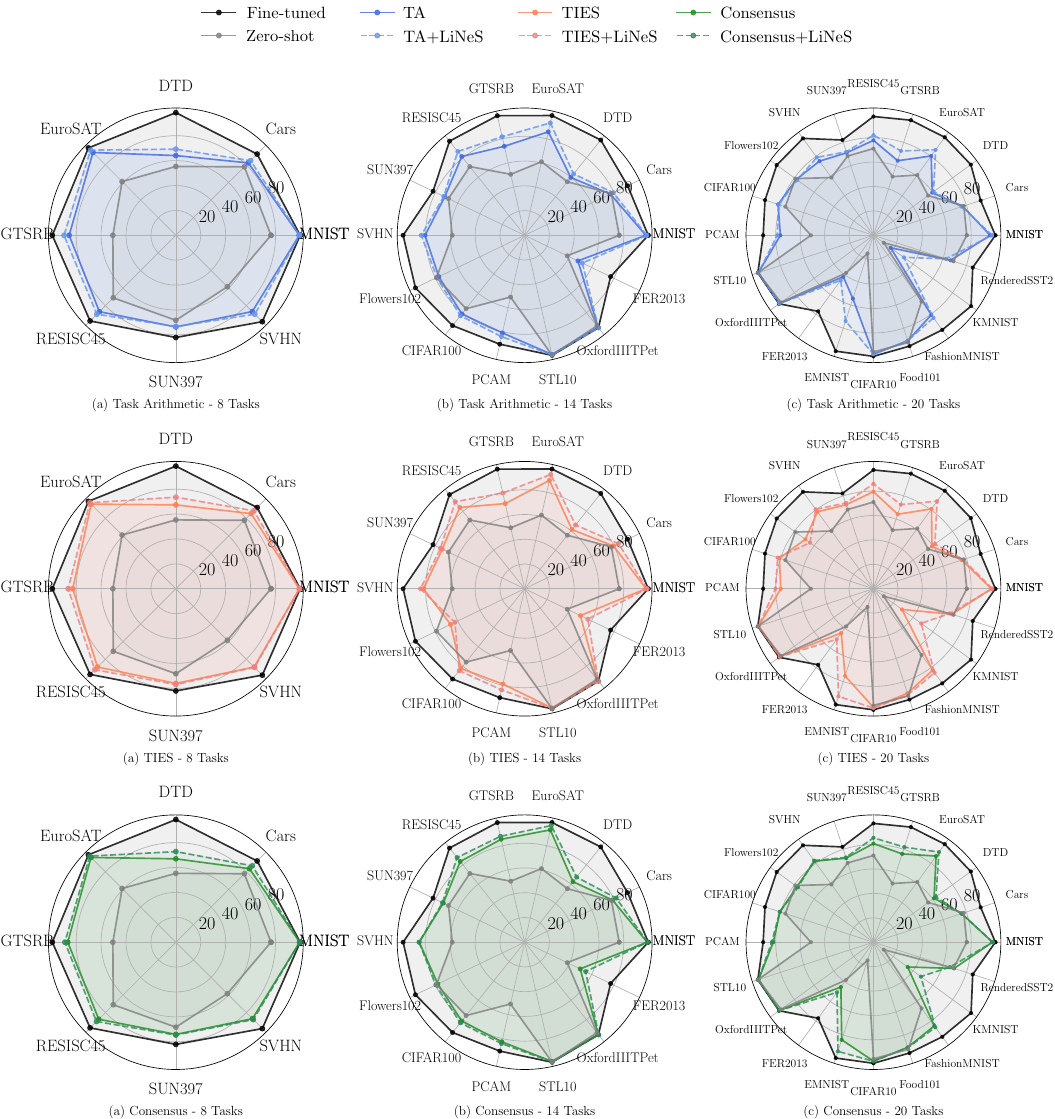}
    \caption{Single-task accuracies for multi-task merging on image classification benchmarks for ViT-L/14.}
    \label{fig:ind_vit_l_14}
\end{figure}

The single-task performance is presented in Figure~\ref{fig:ind_vit_b_32} for ViT-B/32, Figure~\ref{fig:ind_vit_b_16} for ViT-B/16, and Figure~\ref{fig:ind_vit_l_14} for ViT-L/14. From the results we observe that our method demonstrates a noticeable improvement over baseline merging techniques across individual tasks in all test scenarios.

\paragraph{Natural Language Processing}

\begin{figure}[t]
    \centering
    \includegraphics[width=1\linewidth]{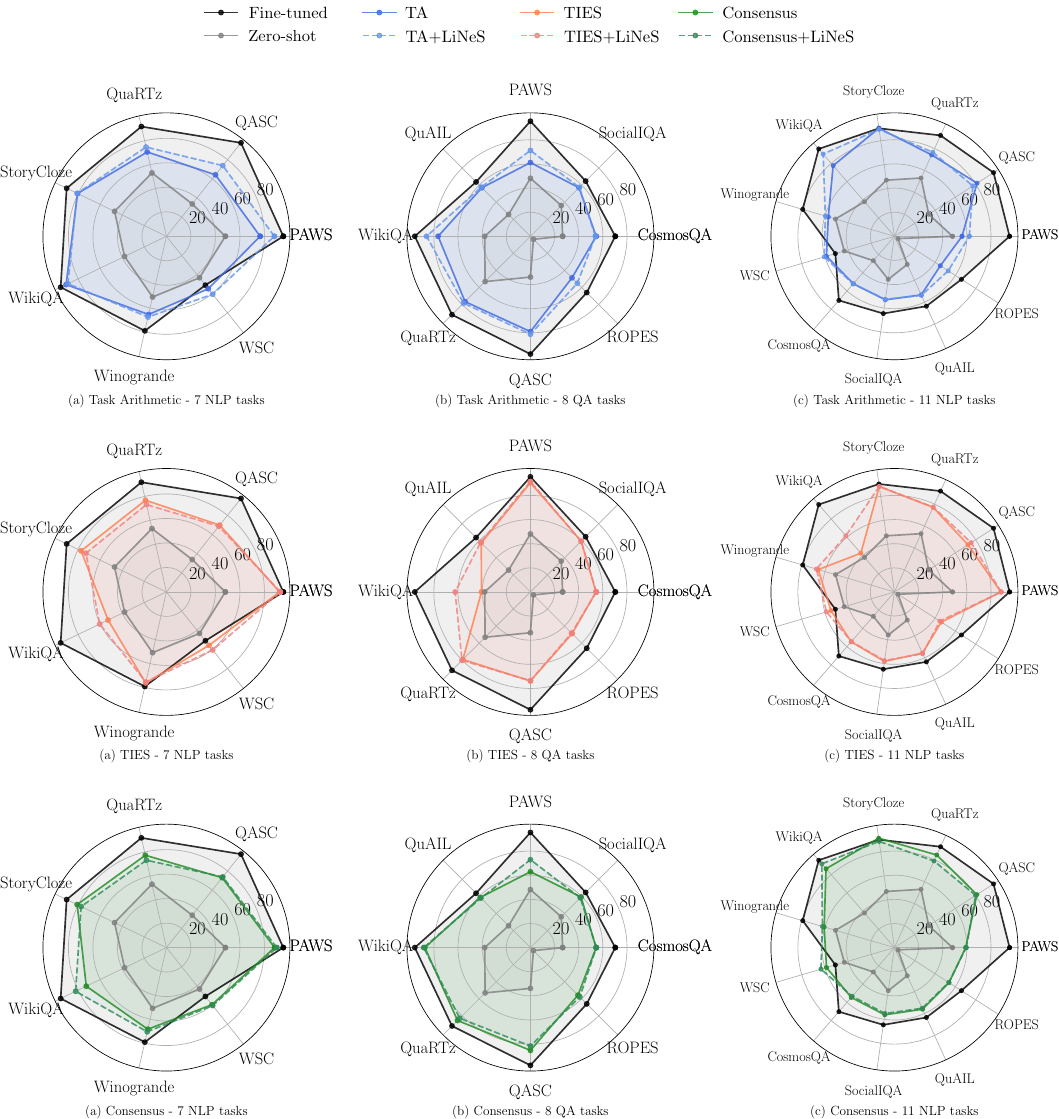}
    \caption{Single-task accuracies for multi-task merging on NLP benchmarks for T5-large.}
    \label{fig:ind_nlp_t5}
\end{figure}

We provide in Figure~\ref{fig:ind_nlp_t5} the detailed single-task performance for the three NLP benchmarks using T5-large, complementary to the results in Table~\ref{tab:nlp with t5}. Similar to the observation in vision, our method provides a consistent improvement over baselines across individual tasks.

\end{document}